\documentclass[a4paper,fleqn]{cas-sc}

\usepackage[numbers,sort&compress]{natbib}
\usepackage{afterpage}     
\usepackage{placeins}      

\def\tsc#1{\csdef{#1}{\textsc{\lowercase{#1}}\xspace}}
\tsc{WGM}
\tsc{QE}

\begin{document}
\let\WriteBookmarks\relax
\def\floatpagepagefraction{1}
\def\textpagefraction{.001}

\shorttitle{CLUE: Controllable Latent space of Unprompted Embeddings}

\shortauthors{K. Park et~al.}  

\title [mode = title]{CLUE: Controllable Latent space of Unprompted Embeddings 
                        for Diversity Management in Text-to-Image Synthesis}

\author[1]{Keunwoo Park}[orcid=0000-0002-1098-9081]
\ead{knupark08@gmail.com}
\credit{Conceptualization, Methodology, Software, Visualization, Writing – original draft}

\affiliation[1]{organization={Departments of Convergence Medicine, 
                University of Ulsan College of Medicine, Asan Medical Center},
                addressline={88, Olympic-ro 43-gil, Songpa-gu}, 
                city={Seoul},
                postcode={05505}, 
                country={Korea}}

\author[1]{Jihye Chae}
\credit{Validation}

\author[2]{Joong Ho Ahn}[orcid=0000-0001-6726-8894]
\ead{meniere@amc.seoul.kr}
\credit{Data curation, Supervision, Validation}

\affiliation[2]{organization={Department of Otorhinolaryngology-Head and Neck Surgery, 
                University of Ulsan College of Medicine, Asan Medical Center},
                addressline={88, Olympic-ro 43-gil, Songpa-gu}, 
                city={Seoul},
                postcode={05505}, 
                country={Korea}}

\author[1]{Jihoon Kweon}[orcid=0000-0003-1831-7615]
\cormark[1]
\ead{kjihoon2@gmail.com}
\credit{Project administration, Supervision, Writing – review and editing}

\cortext[1]{Corresponding author}

\begin{abstract}
Text-to-image synthesis models require the ability to generate diverse images while maintaining stability. To overcome this challenge, a number of methods have been proposed, including the collection of prompt-image datasets and the integration of additional data modalities during training. Although these methods have shown promising results in general domains, they face limitations when applied to specialized fields such as medicine, where only limited types and insufficient amounts of data are available. 
We present CLUE (Controllable Latent space of Unprompted Embeddings), a generative model framework that achieves diverse generation while maintaining stability through fixed-format prompts without requiring any additional data. Based on the Stable Diffusion architecture, CLUE employs a Style Encoder that processes images and prompts to generate style embeddings, which are subsequently fed into a new second attention layer of the U-Net architecture. Through Kullback-Leibler divergence, the latent space achieves continuous representation of image features within Gaussian regions, independent of prompts. 
Performance was assessed on otitis media dataset. CLUE reduced FID to 9.30 (vs. 46.81) and improved recall to 70.29\% (vs. 49.60\%). A classifier trained on synthetic-only data at 1000\% scale achieved an $F_1$ score of 83.21\% (vs. 73.83\%). Combining synthetic data with equal amounts of real data achieved an $F_1$ score of 94.76\%, higher than when using only real data. On an external dataset, synthetic-only training achieved an $F_1$ score of 76.77\% (vs. 60.61\%) at 1000\% scale. The combined approach achieved an $F_1$ score of 85.78\%, higher than when using only the internal dataset. These results demonstrate that CLUE enables diverse yet stable image generation from limited datasets and serves as an effective data augmentation method for domain-specific applications.
\end{abstract}




\maketitle

\section{Introduction}
Generative deep learning has achieved remarkable progress in recent years and has attracted significant interest in various application domains \citep{yang2023diffusion}. Of these advances, text-to-image synthesis—the generation of visually coherent and semantically consistent images from textual descriptions—stands out as both compelling and challenging \citep{tan2023recent,habib2024exploring}. Recent models such as DALL·E \citep{ramesh2021zero,marcus2022very}, Imagen \citep{saharia2022photorealistic} and Stable Diffusion \citep{rombach2022high,esser2024scaling} demonstrate impressive general purpose image generation capabilities. They leverage large scale image-prompt paired datasets to produce visuals that are both diverse and high quality.

While these models have shown remarkable performance, adapting them to specialized domains remains challenging. Fine-tuning is often required to align the model with domain-specific data distributions and stylistic conventions. This requirement is particularly pronounced in medical imaging, where applications demand not only high visual fidelity but also robust stability and precision due to clinical decision-making.

\begin{figure}[t]
	\centering
	\includegraphics[width=\columnwidth]{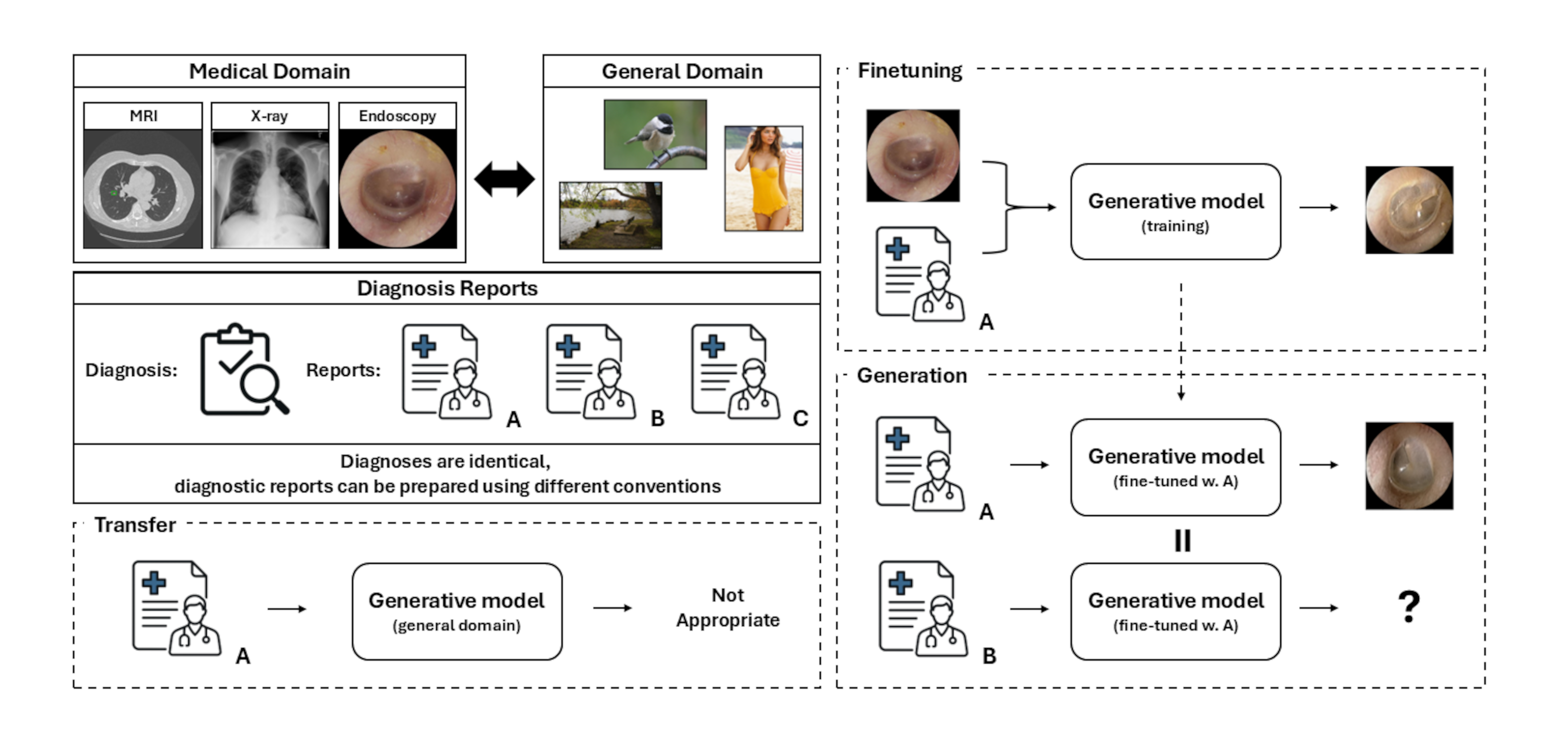}
    \caption{
    Left: Key characteristics of medical datasets. 
    (1) There are notable visual differences between medical and general domain images. 
    (2) Even for the same diagnosis, clinical reports A, B, and C may use different phrasing or expression styles. 
    (3) A model trained on general domain images is not directly applicable to medical image generation.  
    Right: Uncertainty in generative models. 
    When fine-tuned on report A, the model generates accurate images with A-style descriptions, but may produce inappropriate outputs when encountering an unfamiliar phrasing style such as report B.
    }
	\label{domainfig}
\end{figure}

As illustrated in Figure~\ref{domainfig}, the medical domain presents distinct challenges for text-to-image synthesis. First, there exists a significant visual and structural gap between medical and general domain images \citep{yoon2024domain}. Second, diagnostic reports, although clinically equivalent, are written using heterogeneous phrasing conventions, which introduces semantic variability that can confuse generative models. While fine-tuning a model on reports written in one convention (e.g., Report A) allows accurate generation for similar inputs, the model may fail to generalize when exposed to unseen prompts (e.g., Report B), resulting in unpredictable outputs. These constraints consequently render both the diversity and precise control of generated images particularly challenging when image synthesis is conducted solely through the use of identical prompts.

Instability is rooted in the text-to-image models; their generative behavior is tightly coupled to the prompt distribution observed during training. Much of the domain knowledge is implicitly encoded in the training data and thus remains inaccessible through prompts alone. This issue becomes especially problematic in clinical settings, where precise generation is crucial, yet available training data is scarce. Recent efforts to enhance controllability, such as GLIDE \citep{nichol2021improved}, IP-Adapter \citep{ye2023ip}, ControlNet \citep{zhang2023adding} and T2I-Adapter \citep{mou2024t2i}, introduce auxiliary conditioning through inputs like reference images, pose maps or edge maps. While effective in improving generation control, these methods depend on additional data modalities that may not be feasible in low resource or privacy sensitive medical environments.

To address these challenges, we introduce CLUE (Controllable Latent space of Unprompted Embeddings), a novel framework designed to achieve both visual diversity and generation stability without requiring extra data. CLUE augments the Stable Diffusion architecture with a Style Encoder that learns visual attributes from paired images and prompts. These style embeddings are injected into a new second attention layer of the U-Net \citep{ronneberger2015u}, enabling the model to disentangle and control implicit features not captured in the prompt. Kullback-Leibler (KL) divergence \citep{kullback1951information} constraints are applied to enforce a Gaussian distribution in the latent space, facilitating smooth and consistent variation in generated images.

We validate CLUE on an Asan Medical Center (AMC) otitis media dataset comprising images of normal tympanic membrane, otitis media with effusion (OME), and chronic otitis media (COM). Our model achieves a substantial improvement in generation quality, reducing the Fréchet Inception Distance (FID) score from 46.81 to 9.30 and improving recall from 49.60\% to 70.29\%. Synthetic images generated by CLUE, when used as training data, significantly improve downstream classifier performance on the real data, achieving an $F_1$ score of 83.21\%, compared to 73.83\% with reference synthetic data. Combining 100\% scale synthetic data with equal amounts of real data further achieved 94.76\%. For evaluate external generalization, we additionally tested our framework on the AI-HUB otoscopic image dataset. Even on this unseen dataset, models trained exclusively on CLUE-generated synthetic data achieved an $F_1$ score of 72.37\% (vs. 54.26\%), demonstrating robust generalization beyond the institution. When combining 100\% scale synthetic data with equal amounts of real data, CLUE achieved 85.78\% $F_1$ on the AI-HUB dataset, surpassing the 83.76\% achieved by a model trained with only the AMC dataset.

\section{Related Work}
The proposed CLUE framework integrates the complementary strengths of variational autoencoders \citep{kingma2013auto} and diffusion models to improve both fidelity and control diversity in the image synthesis. In this section, we present a comprehensive review of the theoretical foundations and key principles underlying these generative modeling paradigms.

\subsection{Variational Autoencoders}
Variational Autoencoders (VAEs) offer a probabilistic framework for learning a compact latent representation $z \in \mathbb{R}^d$ of high-dimensional data $x$. The encoder $q_\varphi(z|x)$ maps each input $x$ to a Gaussian posterior over latent codes, and the decoder $p_\theta(x|z)$ reconstructs $x$ from a sampled $z$. Training maximizes the evidence lower bound (ELBO) on the marginal log-likelihood $\log p_\theta(x)$, striking a balance between reconstruction fidelity and conformity to a simple prior $p(z)$. To allow gradient based optimization through the stochastic encoder, the reparameterization trick expresses $z$ as a deterministic function of $x$ and an auxiliary noise variable $\epsilon$. Consequently, the trained model can generate new sample images that closely resemble the training data by sampling from a standard Gaussian prior.

Conditional VAEs (CVAEs) \citep{sohn2015learning} extend the VAE framework by introducing an external conditioning variable $c$ (e.g., class labels or text) into both the encoder and the decoder. It enables the generation of images that satisfy specified conditions within the domain on which the model was trained.
\begin{equation}
p_\theta(x|c) = \int p_\theta(x|z,c) p_\theta(z|c) dz
\end{equation}
This conditional likelihood is maximized by optimizing the following evidence lower bound (ELBO):
\begin{equation}
\log p_\theta(x|c) \geq \mathbb{E}_{z \sim q_\phi(z|x,c)} \left[ \log p_\theta(x|z,c) \right] - D_{\text{KL}}(q_\phi(z|x,c) \| p_\theta(z|c))
\end{equation}
The approximate posterior is reparametrized using Gaussian statistics:
\begin{equation}
q_\phi(z|x,c) = \mathcal{N}(\mu_\phi(x,c), \mathrm{diag}(\sigma_\phi(x,c)^2)), \quad
z = \mu_\phi(x,c) + \sigma_\phi(x,c) \odot \epsilon, \quad \epsilon \sim \mathcal{N}(0,I)
\end{equation}

where $\odot$ denotes element-wise multiplication. This conditioning enables more controlled data synthesis, making CVAEs particularly useful in scenarios where generation needs to be steered by external signals, such as text-to-image tasks.

\subsection{Diffusion Models}
Diffusion models define a forward noising process that gradually corrupts data and a learned reverse process that recovers clean samples \citep{sohl2015deep,nichol2021improved}. This framework unites the stability of probabilistic modeling with the ability to produce high-quality samples, and has achieved state-of-the-art results across image, audio, and text domains.

$x_t$ represents the noisy data at timestep $t$, derived from the original clean data $x_0$ through a forward process that gradually adds Gaussian noise. $\bar{\alpha}_t$ controls the cumulative noise schedule and $\beta_t$ defines the noise schedule parameters added at each timestep. The noise term $\epsilon \sim \mathcal{N}(0,I)$ serves as the primary training target in the simplified objective $\mathcal{L} = \mathbb{E}[||\epsilon - \epsilon_\theta(x_t,t)||^2]$.

In addition to the common noise prediction and clean sample parameterization, diffusion models may be trained to predict a velocity term $v_t$, which combines the clean signal and noise in a single target \citep{salimans2022progressive}. This v-prediction parameterization has been shown to yield improved numerical stability for few step sampling scenarios.
\begin{equation}
x_t = \sqrt{\bar{\alpha}_t} x_0 + \sqrt{1-\bar{\alpha}_t} \epsilon, \quad
v_t = \sqrt{\bar{\alpha}_t} \epsilon - \sqrt{1-\bar{\alpha}_t} x_0, \quad
\bar{\alpha}_t = \prod_{i=1}^{t}(1-\beta_i) 
\end{equation}

The relationships between these parameterizations enable flexible conversion:
\begin{equation}
\hat{v} = v_\theta(x_t, t, c), \quad
\hat{x}_0^t = \sqrt{\bar{\alpha}_t} x_t - \sqrt{1-\bar{\alpha}_t} \hat{v}, \quad
\hat{\epsilon}^t = \sqrt{1-\bar{\alpha}_t} x_t + \sqrt{\bar{\alpha}_t} \hat{v}
\end{equation}

$v_\theta$ is the neural network predicting velocity, $c$ is the conditioning information, and $\hat{x}_0^t, \hat{\epsilon}^t$ are the estimated clean image and noise at timestep $t$, respectively.
\begin{equation}
x_{t-1} = \sqrt{\bar{\alpha}_{t-1}} \hat{x}_0^t + \sqrt{1-\bar{\alpha}_{t-1}} \hat{\epsilon}^t 
\end{equation}

Substituting the conversion formulas:
\begin{equation}
x_{t-1} = \sqrt{\bar{\alpha}_{t-1}} \left( \sqrt{\bar{\alpha}_t} x_t - \sqrt{1-\bar{\alpha}_t} \hat{v} \right) 
+ \sqrt{1-\bar{\alpha}_{t-1}} \left( \sqrt{1-\bar{\alpha}_t} x_t + \sqrt{\bar{\alpha}_t} \hat{v} \right)
\end{equation}

This reverse process enables sampling by iteratively denoising from $x_T \sim \mathcal{N}(0,I)$ to $x_0$. Conditional diffusion models extend this framework by incorporating additional information to guide the reverse process through the conditioning variable $c$ in $v_\theta(x_t,t,c)$.
\begin{equation}
\mathcal{L}_t = ||v_t - v_\theta(x_t, t, c)||^2 
= ||\sqrt{\bar{\alpha}_t} \epsilon - \sqrt{1-\bar{\alpha}_t} x_0 - v_\theta(x_t, t, c)||^2
\end{equation}

Latent diffusion models (LDMs) operate in a lower dimensional latent space obtained via an autoencoder, substantially reducing computational cost while preserving semantic structure. Stable Diffusion exemplifies this approach by integrating powerful text encoders with efficient latent-space diffusion, achieving photorealistic synthesis from natural language prompts with high fidelity and diversity.

\section{Method}
We introduce CLUE (Controllable Latent of Unprompted Embedding), a novel architecture that enhances Stable Diffusion's controllability by learning prompt disentangled visual representations. Our approach builds upon the Stable Diffusion 2.1 (SD 2.1) U-Net backbone, treating its forward diffusion process and a style encoder as the CVAE encoder and the denoising U-Net as the CVAE decoder.
\begin{figure}[h]
    \centering
    \includegraphics[width=\columnwidth]{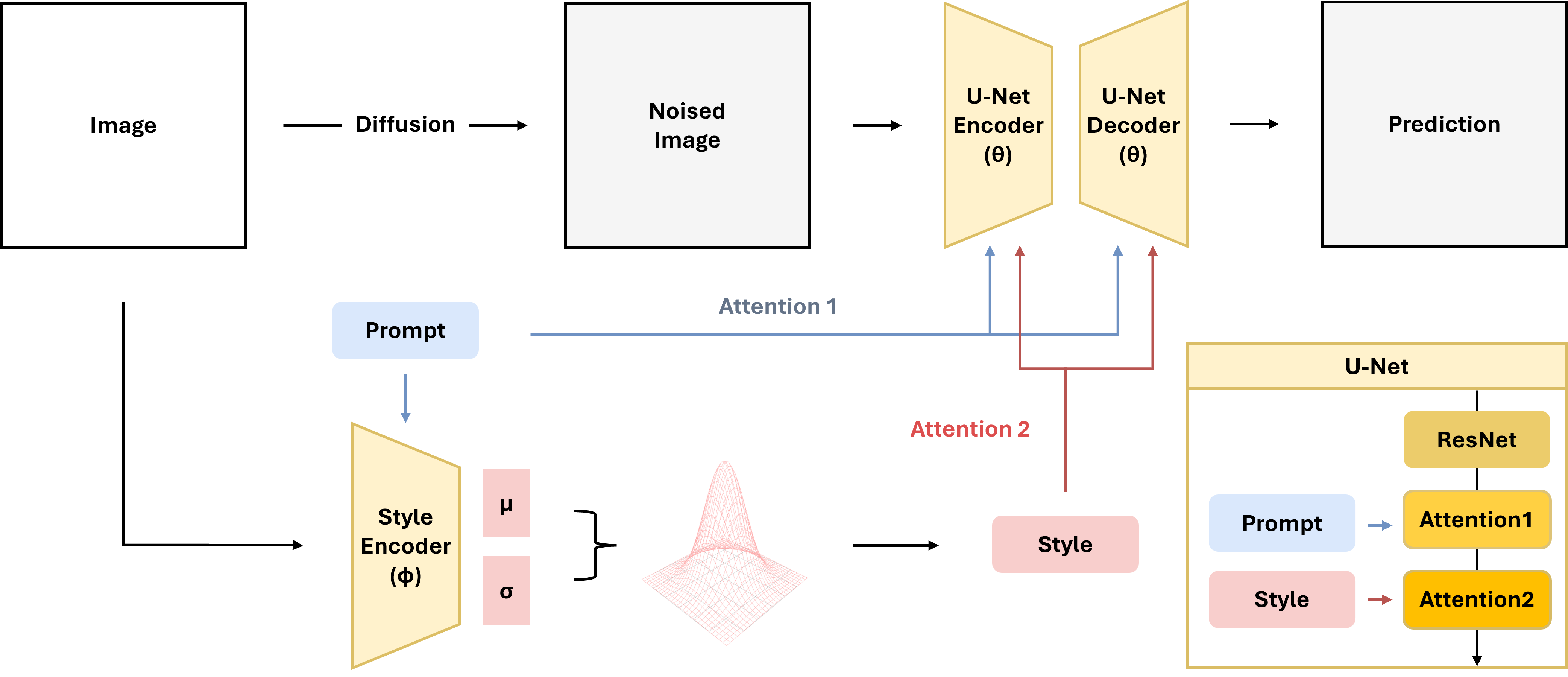}
    \caption{Structure of CLUE. The noised image passed through the U-Net with the prompt to predict a velocity. The clean image and prompt are sent into the style encoder to parameterize a Gaussian posterior from which a style latent $s \in \mathbb{R}^{1024}$ is sampled.}
    \label{fig:clue_structure}
\end{figure}

\subsection{Style Encoding and Sampling}
Specifically, we reuse the SD 2.1 U-Net encoder architecture in our CLUE style encoder, appending only a single fully connected layer at the final feature map to predict the mean $\mu_\phi(x,c)$ and standard deviation $\sigma_\phi(x,c)$ of Gaussian style latent. This style latent is injected into every down- and up-sampling block of the U-Net via dedicated cross-attention layers—one per block—whose key/value projections are derived from the sampled style (Figure~\ref{fig:clue_structure}).

\subsection{Double Cross-Attention Mechanism}
For image reconstruction, the U-Net denoiser takes as inputs the noised image and a timestep embedding, and conditions on both the prompt and the sampled style latent. Concretely, within each U-Net block the features are first modulated by a cross-attention layer attending to the prompt embedding, and then by a second cross-attention layer attending to the style latent.
\begin{equation}
\text{Attention}(Q,K,V) = \text{softmax}\left(\frac{QK^\top}{\sqrt{d_k}}\right)V
\end{equation}
\begin{equation}
z' = \text{Attention}(zW_Q^1, cW_K^1, cW_V^1), \quad
z'' = \text{Attention}(z'W_Q^2, sW_K^2, sW_V^2)
\end{equation}

$z$ represents the latent features, $c$ is the text embedding from CLIP \citep{radford2021learning} encoder, $s \in \mathbb{R}^{1024}$ is the sampled style latent, $d_k$ is the key dimension, and $W_Q$, $W_K$, $W_V$ are learned projection matrices. This dual cross-attention mechanism enables both semantic alignment through text and stylistic control through the learned style representation.

\subsection{Training Objective}

Loss $\mathcal{L}(\theta,\phi)$ comprises two terms. The first term, the denoising loss, is defined as the expected mean squared error between the model's predicted velocity $v_\theta(x_t,t,c,s)$ and the true velocity target $v_t$. This expectation is taken over the data distribution of original images $x$ and text conditions $c$, a sampled timestep $t$, standard Gaussian noise $\epsilon$ used to produce the noised image $x_t$, and the style latent $s$ sampled from the encoder $SE_\phi$. The second term, the KL regularization loss, computes the Kullback-Leibler divergence between the approximate posterior $SE_\phi(s|x,c)$ and the standard Gaussian prior $\mathcal{N}(0,I)$.
\begin{equation}
\mathcal{L}(\theta,\phi) = \underbrace{\mathbb{E}_{x,c,t,\epsilon \sim \mathcal{N}(0,I), s \sim SE_\phi(s|x,c)} \left[ ||v_t - v_\theta(x_t,t,c,s)||_2^2 \right]}_{\mathcal{L}_{\text{denoise}}(\theta,\phi)} 
+ \lambda \underbrace{\mathbb{E}_{x,c} \left[ D_{\text{KL}}(SE_\phi(s|x,c) \| \mathcal{N}(0,I)) \right]}_{\mathcal{L}_{\text{KL}}(\phi)}
\end{equation}

where $\lambda$ is the KL regularization weight. This objective combines the benefits of diffusion model training with variational autoencoder regularization, enabling controllable generation through the learned style space.
$SE_\phi$ denotes the style encoder parameterized by $\phi$. The reparameterization trick enables gradient-based optimization through the stochastic sampling process.
\begin{equation}
SE_\phi(s|x,c) = \mathcal{N}(\mu_\phi(x,c), \text{diag}(\sigma_\phi(x,c)^2)), \quad
s = \mu_\phi(x,c) + \sigma_\phi(x,c) \odot \epsilon_s, \quad \epsilon_s \sim \mathcal{N}(0,I)
\end{equation}


\section{Experiment}
\subsection{Environment}
All experiments were carried out on a dedicated compute node equipped with Intel Xeon Gold 6458Q CPUs and sixteen 64 GB DDR4 memory modules (total system memory: 1,024 GB). Accelerated sampling and training were performed on eight NVIDIA L40S GPUs (48 GB each), running NVIDIA driver 565.57.01 with CUDA 12.6. Our implementation is based on Python 3.11.11. Core model training and inference leverage PyTorch 2.6.0+cu126. We build upon the Hugging Face Diffusers framework (0.33.0.dev0) for diffusion‐based components and employ the timm library (0.4.12) \citep{rw2019timm} for vision backbones and utility routines.

\subsection{Dataset}
Asan Medical Center (AMC) endoscopic dataset is retrospectively collected from patients who visited the Department of Otolaryngology at Seoul Asan Medical Center between January 2018 and December 2020. All video sequences were originally acquired for clinical diagnostic purposes; from each sequence, a single frame exhibiting the entire tympanic membrane was anonymized and exported for analysis. This study adhered to the Declaration of Helsinki and was approved with a waiver of informed consent by the Institutional Review Board of Seoul Asan Medical Center (IRB No. 2021-0837). Two otolaryngology specialists (one with 26 years' experience, the other with 5 years) independently and blindly reviewed each anonymized image. Discrepancies were resolved by consensus.

Images were assigned to one of three categories. The normal class denotes a healthy, disease free normal tympanic membrane, characterized by an intact, translucent appearance without middle ear effusion or perforation. Otitis media with effusion (OME) is characterized by the presence of fluid within the middle ear cavity behind an intact tympanic membrane, often manifesting as an air-fluid level or a diffuse amber-hued effusion visible through the membrane. Chronic otitis media (COM) is defined by a perforation of the tympanic membrane, regardless of perforation size or location. We collected 1,500 images for each of the normal, OME, and COM categories from Seoul Asan Medical Center, ensuring balanced representation across all classes. Samples of the dataset are shown in Appendix~\ref{sec:appendixA}.
\begin{table}[h]
    \centering
    \caption{AMC endoscopic dataset composition by class, prompt, and split}
    \label{tbl:dataset_composition}
    \small
    \begin{tabular*}{\columnwidth}{@{\extracolsep{\fill}}llccc@{}}
        \toprule
        & & \multicolumn{2}{c}{Dataset 1} & \\
        \cmidrule(lr){3-4}
        Class & Prompt & A & B & Dataset 2 \\
        \midrule
        Normal & Tympanic membrane is translucent, no fluid, and no perforation.        & 900 & 300 & 300 \\
        \addlinespace
        OME    & Tympanic membrane has fluid behind it, often with a dull or amber hue. & 900 & 300 & 300 \\
        \addlinespace
        COM    & Tympanic membrane is perforated. & 900 & 300 & 300                                       \\
        \bottomrule
    \end{tabular*}
\end{table}

\begin{table}[h]
    \centering
    \caption{AI-HUB dataset composition before and after filtering}
    \label{tbl:aihub_preprocessing}
    \small
    \begin{tabular*}{\columnwidth}{@{\extracolsep{\fill}}lccc@{}}
        \toprule
        Class & Before filtering & After filtering & Random selection \\
        \midrule
        Normal & 13,502 & 11,519 & 3,000 \\
        \addlinespace
        OME & 5,856 & 5,211 & 3,000 \\
        \addlinespace
        COM & 3,303 & 3,043 & 3,000 \\
        \midrule
        Total & 22,661 & 19,773 & 9,000 \\
        \bottomrule
    \end{tabular*}
\end{table}

To assess model performance, the data was partitioned into two subsets. Dataset 1 was used to train the generative model; within this dataset, subset A served as the classification model's training set (for evaluating the generative model's performance), while subset B was used for validation. Dataset 2, being entirely independent of both generative and classification model training, was reserved as the test set. Table~\ref{tbl:dataset_composition} summarizes the number of samples per class alongside their corresponding textual prompts.

To enhance the robustness and generalizability of our model, we additionally utilized the publicly available AI-HUB otoscopic image dataset. This dataset comprises 49,614 disease-related images across seven categories. The external data underwent filtering to standardize the endoscopic view. Only images with detectable endoscopic circles were retained. From this preprocessed dataset, we selected 3,000 images each from the normal, OME, and COM categories for testing purposes.

\subsection{Preprocessing}
Each raw RGB endoscopic frame is first converted to grayscale and binarized with a fixed intensity threshold of 26 to isolate the bright circular field of view, after which the largest contour is extracted and subjected to a Euclidean distance transform, its maximum distance point defining the center and radius of the maximal inscribed circle. A binary mask of this circle is then applied to the original color image to remove all pixels outside the endoscopic aperture, the masked region is cropped via its axis aligned bounding rectangle, and the result is resized to $512 \times 512$ pixels using bilinear interpolation with nearest neighbor resizing for the mask. This procedure yields uniformly centered, fixed size images that are free of peripheral artifacts and optimized for modeling.

\begin{figure}[t]
    \centering
    \includegraphics[width=\columnwidth]{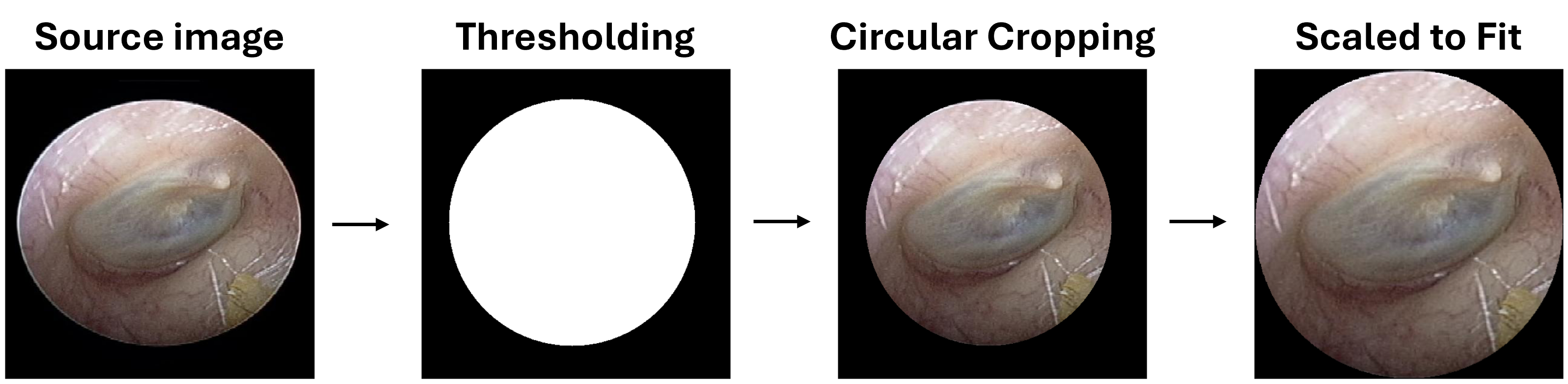}
	\caption{Overview of the four stage preprocessing applied to each endoscopic frame}
	\label{preprocessing}
\end{figure}

\subsection{Training}
Table~\ref{tbl:genparam} summarizes the settings used to train generative models. We built CLUE on the Stable Diffusion 2.1 architecture, augmenting the backbone with 1,024 style embeddings to capture diverse appearances in generated outputs. Denoising was performed with a DDIM scheduler \citep{song2020denoising} over 1,000 timesteps following a linear beta schedule and employing the 'v-prediction' parameterization. All input images were resized to $512 \times 512 \times 3$ pixels and normalized to the $[-1,+1]$ range. Training was conducted in full FP32 precision with a mini-batch size of 4, optimizing with AdamW at a constant learning rate of $1 \times 10^{-5}$ for up to 10,000 update steps. Figure~\ref{preprocessing} illustrates the four-stage preprocessing pipeline applied to each endoscopic frame.

Using these exact settings, we trained our proposed CLUE model and, for comparison, fine-tuned a Stable Diffusion 2.1 checkpoint (referred to as the Vanilla) under identical hyperparameters. Learning curves of the models are shown in Appendix~\ref{sec:appendixB}.

\begin{table}[h]
    \centering
    \caption{Training configurations for generative models}\label{tbl:genparam}
    \begin{tabular*}{\columnwidth}{@{\extracolsep{\fill}}ll@{}}
        \toprule
        Parameter & Value \\
        \midrule
        Base model & SD 2.1 \\
        Diffusion scheduler & DDIMScheduler \\
        Diffusion timesteps & 1,000 \\
        Prediction type & v-prediction \\
        Resolution & $512 \times 512 \times 3$ \\
        Normalization & [-1, +1] \\
        Precision & FP32 \\
        Batch & 4 \\
        Optimizer & AdamW \\
        Learning rate & 1e-5 \\
        Learning scheduler & Constant \\
        Maximum steps & 10,000 \\
        \bottomrule
    \end{tabular*}
\end{table}

\subsection{Synthetic Dataset Generation}
\begin{figure}[t]
    \centering
    \includegraphics[width=\columnwidth]{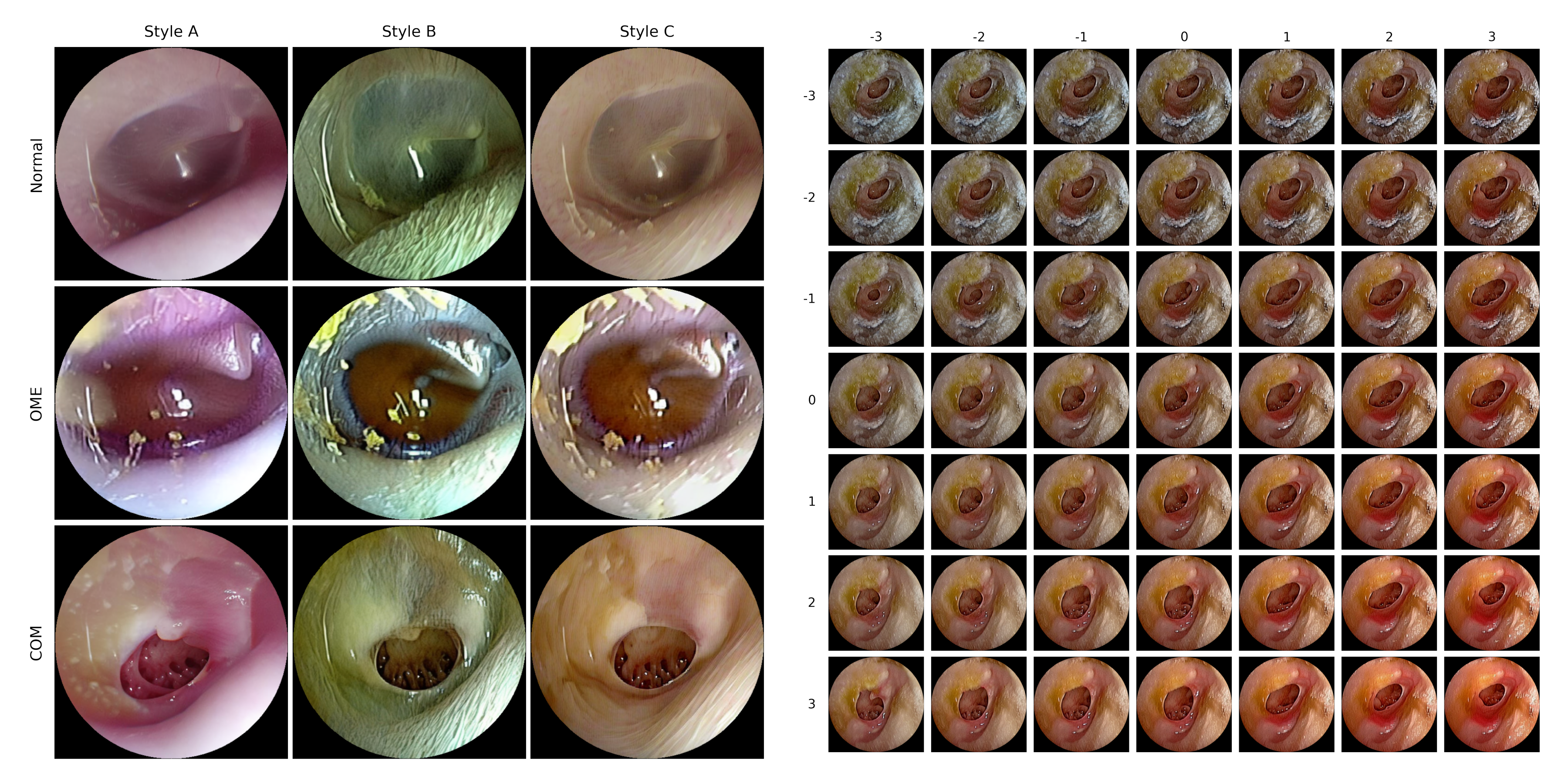}
    \caption{(Right) Examples of generated images by CLUE across different style embeddings. Each row represents a different diagnostic category (Normal, OME, COM), while columns Style A, B, and C demonstrate the sampled style vectors while maintaining the same textual prompt and gaussian noise. (Left) Examples of generated COM images by CLUE. Changing two components of style vector from -3 to 3 and fixed the gaussian noise. Images are continuously changed with style.}\label{fig:samplegen}
\end{figure}

To ensure statistical reliability, we conducted 10 independent experiments, each initialized with a distinct random seed. In each run, we sampled 27,000 Gaussian noise tensors of shape $4 \times 64 \times 64$ and drew corresponding style vectors of dimension $1 \times 1024$ from Gaussian distributions with standard deviations of 0.0, 0.1, 0.3, and 0.5 within the CLUE framework. Each model configuration generated a total of 27,000 synthetic images, evenly balanced across three classes (9,000 images per class).
Overall, this procedure produced 10 independent datasets comprising 270,000 images, enabling robust statistical analyses across all conditions. Generated images were finalized by applying the same preprocessing pipeline used for the original dataset. Table~\ref{tbl:generated_images} details the number of generated images per class for training datasets under the Vanilla and four CLUE configurations, aggregated over the 10 experimental runs. Figure~\ref{fig:samplegen} showcases examples of images generated by CLUE across different style embeddings, demonstrating the framework's ability to produce diverse visual variations while maintaining diagnostic consistency.
\begin{table}[h]
    \centering
    \caption{Number of generated images per class for training datasets across Vanilla and four CLUE configurations. 
    Values represent the number of images per class in each of the 10 independent experiments.}\label{tbl:generated_images}
    \small
    \setlength{\tabcolsep}{3pt}
    \begin{tabular*}{\columnwidth}{@{\extracolsep{\fill}}lcccccccccc@{}}
        \toprule
        & \multicolumn{1}{c}{Vanilla} & \multicolumn{1}{c}{$\sigma=0.0$} & \multicolumn{1}{c}{$\sigma=0.1$} & \multicolumn{1}{c}{$\sigma=0.3$} & \multicolumn{1}{c}{$\sigma=0.5$} \\
        \midrule
        Normal & 9,000 & 9,000 & 9,000 & 9,000 & 9,000 \\
        OME & 9,000 & 9,000 & 9,000 & 9,000 & 9,000 \\
        COM & 9,000 & 9,000 & 9,000 & 9,000 & 9,000 \\
        \midrule
        \multicolumn{11}{l}{\textit{Total images across all 10 experiments: 270,000}}
    \end{tabular*}
\end{table}

\section{Results}
Generated images were evaluated using an InceptionV3 \cite{szegedy2016rethinking} classifier trained from scratch for our domain. Prior to training, all images were resized to $299 \times 299 \times 3$ pixels and normalized to the $[-1,+1]$ range. The model was initialized with random weights (without using pre-trained parameters) and trained in FP32 precision with a batch size of 32. Optimization employed the Adam algorithm with an initial learning rate of $1 \times 10^{-3}$, and a ReduceLROnPlateau scheduler (patience = 10 epochs, reduction factor = 0.5) adjusted the learning rate based on validation loss. Training ran for 300 epochs, with validation metrics recorded at the end of each epoch, using the hyperparameter settings summarized in Table~\ref{tbl:classification_training}. Examples of synthetic datasets are shown in Appendix~\ref{sec:appendixC}.

\begin{table}[h]
    \caption{Training configurations for classification models}\label{tbl:classification_training}
    \begin{tabular*}{\columnwidth}{@{\extracolsep{\fill}}ll}
        \toprule
        Parameter & Value \\
        \midrule
        Base model & InceptionV3 \\
        Resolution & $299 \times 299 \times 3$ \\
        Normalization & [-1, +1] \\
        Precision & FP32 \\
        Batch & 32 \\
        Optimizer & Adam \\
        Learning rate & 1e-3 \\
        Scheduler (patience, factor) & ReduceLROnPlateau (10, 0.5) \\
        Epochs (early stopping patience)  & 300 (30) \\
        \bottomrule
    \end{tabular*}
\end{table}

\subsection{Feature Space Analysis}
Principal Component Analysis (PCA) \citep{pearson1901liii,hotelling1933analysis} was performed using features extracted from an InceptionV3 model fine-tuned on Dataset 1A. This domain adapted model better captures endoscopic image characteristics compared to generic pre-trained models. For each variant, we randomly sampled 1,500 images from each diagnostic category, normal, OME, and COM, yielding 4,500 images per variant.

\begin{figure}[t]
    \centering
    \includegraphics[width=\columnwidth]{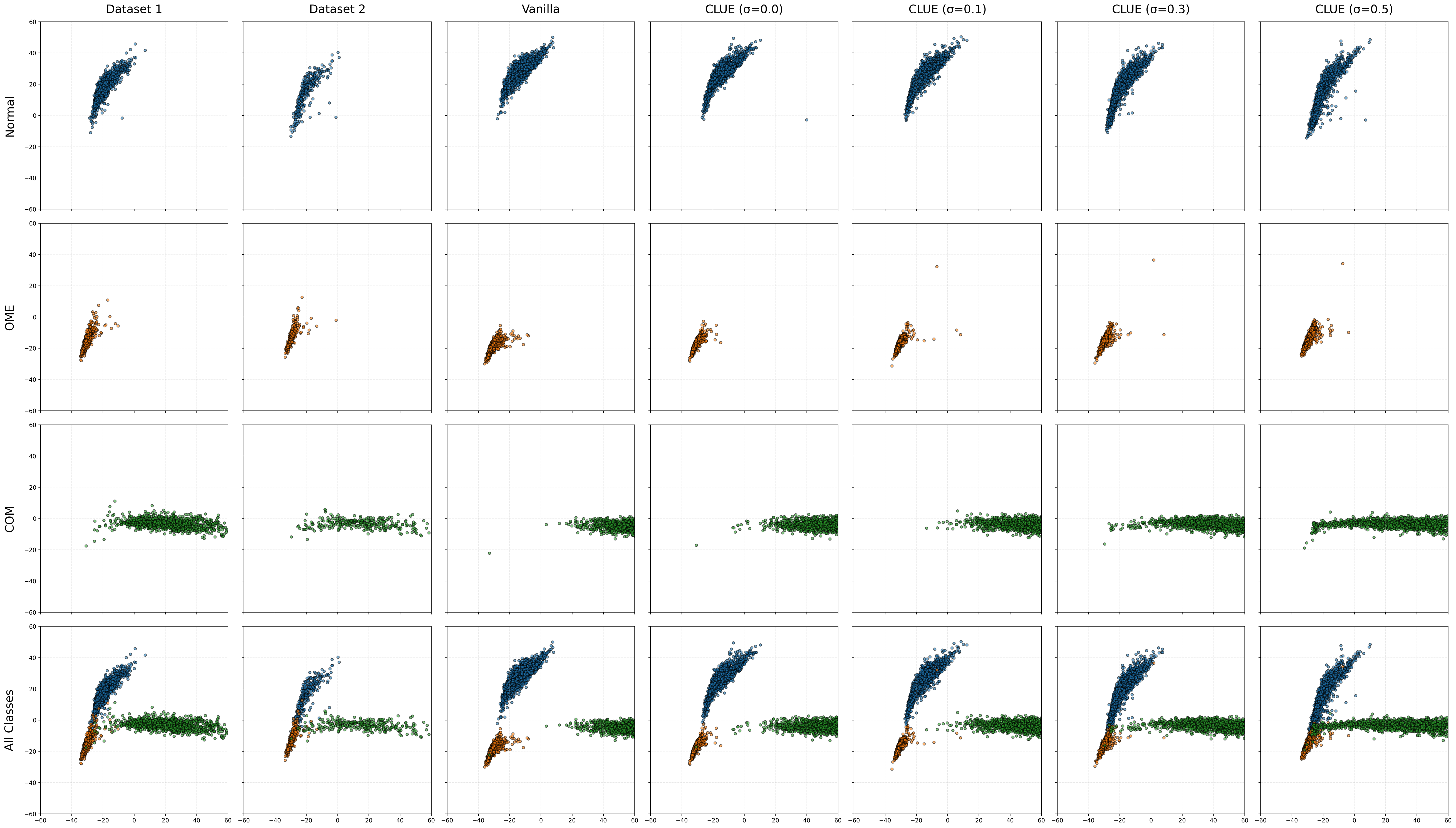}
    \caption{PCA visualization of feature space distributions}\label{fig:pca}
\end{figure}

Figure~\ref{fig:pca} presents the PCA projection of feature distributions across different data sources and model variants. A clear progression is observed as the style variance parameter $\sigma$ increases from 0.0 to 0.5. Vanilla fine-tuning and low variance CLUE ($\sigma=0.0, 0.1$) produce tightly clustered distributions with limited intra-class diversity. As $\sigma$ increases to 0.3 and 0.5, the feature distributions expand while maintaining class specific clustering patterns, suggesting that higher style variance successfully introduces controlled diversity without compromising diagnostic integrity.

Comparing generated images with real data reveals that CLUE with moderate to high style variance ($\sigma=0.3, 0.5$) better approximates the natural feature distribution of real endoscopic images. The vanilla model and low variance CLUE variants generating images that cluster in narrow regions of the feature space. In contrast, higher variance settings produce distributions that more closely match the spread and coverage of real data.

\subsection{Fréchet Inception Distance}
We evaluated the visual quality of generated images using Fréchet Inception Distance (FID) \citep{heusel2017gans}, a widely adopted metric that measures the distance between feature distributions of real and generated images. FID is computed as:
\begin{equation}
\text{FID} = ||\mu_r - \mu_g||^2 + \text{Tr}(\Sigma_r + \Sigma_g - 2(\Sigma_r\Sigma_g)^{1/2})
\end{equation}
where $\mu_r, \Sigma_r$ and $\mu_g, \Sigma_g$ represent the mean and covariance of features extracted from real and generated images, respectively. Lower FID scores indicate higher quality and more realistic generated images. Features were extracted using an InceptionV3 model fine-tuned on Dataset 1A to better capture domain-specific characteristics.

\begin{figure}[t]
    \centering
    \includegraphics[width=\columnwidth]{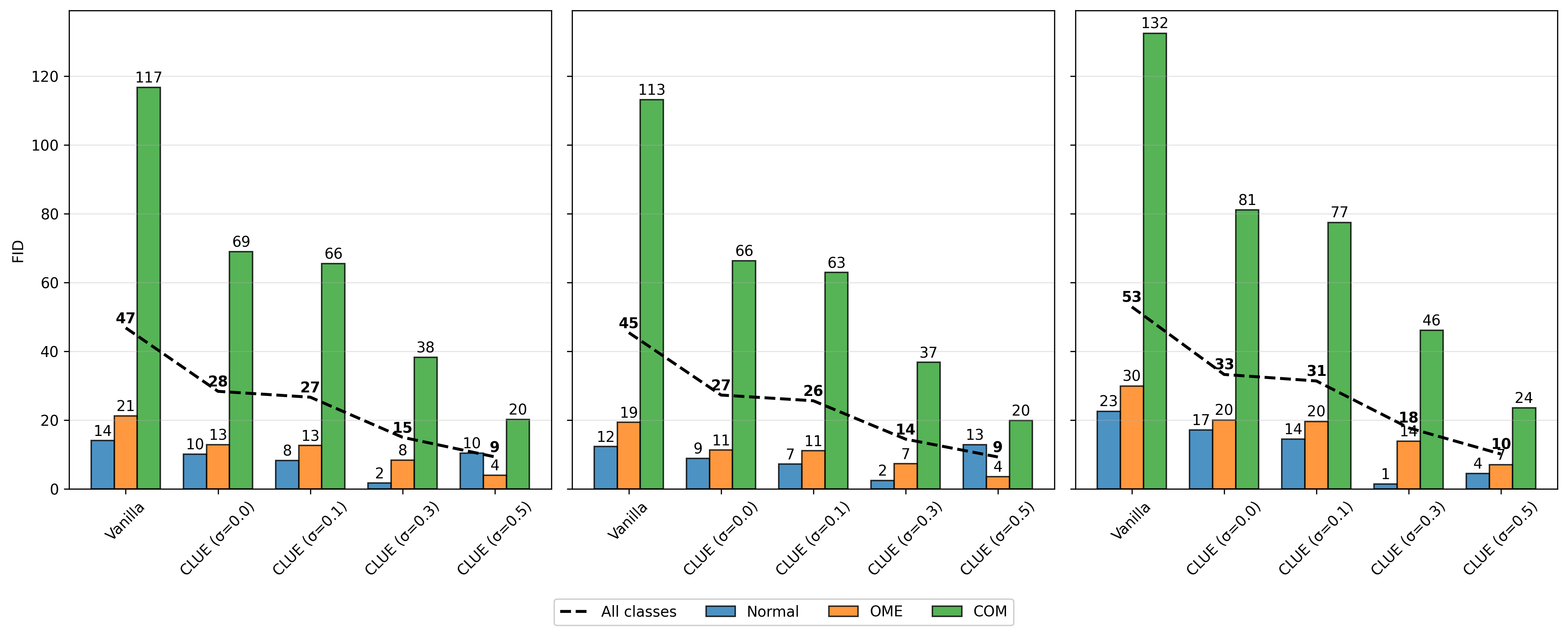}
    \caption{FID scores on (Left) Dataset 1 + 2, (Center) Dataset 1 and (Right) Dataset 2}\label{fig:fid_scored}
\end{figure}

Table~\ref{tbl:fid_scores} and Figure~\ref{fig:fid_scored} present FID scores comparing generated images against real endoscopic images. CLUE consistently outperformed vanilla fine-tuning across all diagnostic categories, with overall FID improving from 46.81 to 9.30 as style variance increased to $\sigma=0.5$. The most substantial improvements were observed for COM images, where FID decreased from 116.77 (vanilla) to 20.18 ($\sigma=0.5$), representing an 82.7\% reduction. This significant improvement for COM suggests that style embeddings are particularly effective at capturing the complex visual patterns of tympanic membrane perforations.

Notably, the performance gap between Dataset 1 and Dataset 2 suggests good generalization, with Dataset 2 showing only slightly higher FID scores despite being completely held out from training. This validates that CLUE learns generalizable style representations rather than overfitting to the training distribution.
\begin{table}[h]
    \caption{FID scores between real and synthetic images}\label{tbl:fid_scores}
    \centering
    \scriptsize
    \setlength{\tabcolsep}{4pt}
    \begin{tabular*}{\columnwidth}{@{\extracolsep{\fill}}lcccccccccccc@{}}
        \toprule
        \multirow{2}{*}{Model}
        & \multicolumn{4}{c}{Dataset 1 + Dataset 2}
        & \multicolumn{4}{c}{Dataset 1}
        & \multicolumn{4}{c}{Dataset 2} \\
        \cmidrule(lr){2-5}\cmidrule(lr){6-9}\cmidrule(lr){10-13}
        & Normal & OME & COM & All
        & Normal & OME & COM & All
        & Normal & OME & COM & All \\
        \midrule
        Vanilla            & 14.03 & 21.23 & 116.77 & 46.81 
                           & 12.34 & 19.33 & 113.18 & 45.42 
                           & 22.56 & 29.92 & 132.48 & 52.91 \\
        CLUE ($\sigma=0.0$)& 10.11 & 12.83 &  68.98 & 28.35 
                           &  8.86 & 11.27 &  66.35 & 27.27 
                           & 17.13 & 20.00 &  81.08 & 33.26 \\
        CLUE ($\sigma=0.1$)&  8.23 & 12.58 &  65.52 & 26.65 
                           &  7.21 & 11.08 &  62.91 & 25.61 
                           & 14.44 & 19.59 &  77.49 & 31.37 \\
        CLUE ($\sigma=0.3$)& \textbf{1.67} &  8.35 &  38.27 & 14.99 
                           & \textbf{2.47} &  7.28 &  36.78 & 14.45 
                           & \textbf{1.44} & 13.85 &  46.07 & 17.73 \\
        CLUE ($\sigma=0.5$)& 10.39 & \textbf{3.97} & \textbf{20.18} & \textbf{9.30} 
                           & 12.83 & \textbf{3.57} & \textbf{19.88} & \textbf{9.26} 
                           &  4.45 & \textbf{6.97} & \textbf{23.50} & \textbf{10.11} \\
        \bottomrule
    \end{tabular*}
\end{table}

\subsection{$k$-NN Recall Performance}
To assess the semantic consistency of generated images, we evaluated how well synthetic images preserve the feature representations of their corresponding real images. We computed the recall performance using $k$-nearest neighbor ($k$-NN) distances \citep{cover1967nearest} in the feature space extracted from the same InceptionV3 model fine-tuned on Dataset 1A (as used for FID computation), with $k=10$, where recall measures the proportion of real samples that have at least one generated sample within a threshold distance.

The recall metric is defined as:
\begin{align}
\text{Recall} &= \frac{1}{|R|} \sum_{r \in R} \mathbb{1}[d(r, \text{NN}_G(r)) \leq \tau]
\end{align}
where $R$ and $G$ denote the sets of real and generated images, respectively; $\text{NN}_G(r)$ is the $k$-th nearest neighbor among generated images to a real image $r$; $d(\cdot,\cdot)$ is the Euclidean distance in feature space; and $\tau$ is an adaptive threshold set to the median of all $k$-NN distances.

\begin{figure}[t]
    \centering
    \includegraphics[width=\columnwidth]{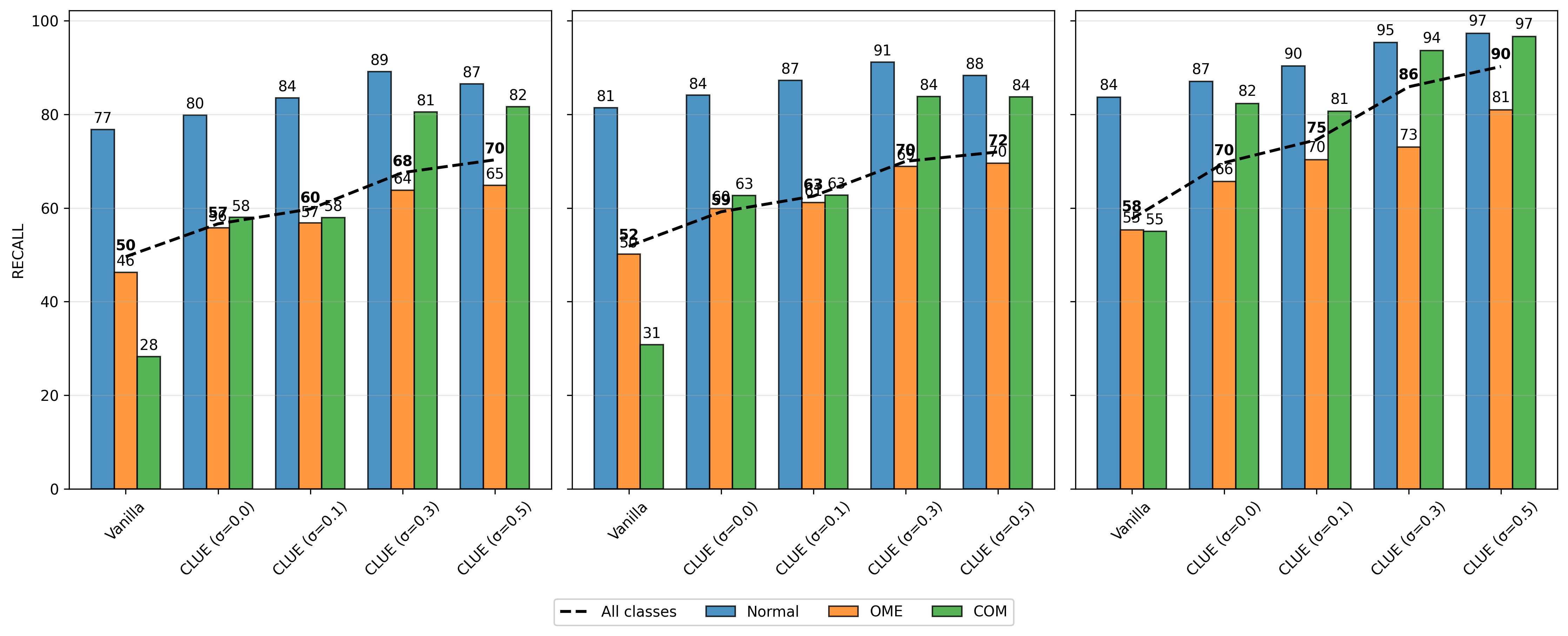}
    \caption{Recall scores on (Left): Dataset 1 + 2, (Center): Dataset 1 and (Right): Dataset 2}\label{fig:recall}
\end{figure}

Table~\ref{tbl:recall} and Figure~\ref{fig:recall} report recall scores for retrieving real images via their nearest synthetic counterparts. Aggregating across all classes, recall rose markedly from 49.60\% (vanilla) to 67.56\% for CLUE with $\sigma=0.3$ and further to 70.29\% for CLUE with $\sigma=0.5$. Breaking this down by class, normal improved from 76.80\% (vanilla) to 89.13\% (CLUE, $\sigma=0.3$) and 86.53\% (CLUE, $\sigma=0.5$), OME from 46.27\% to 63.80\% and 64.87\%, and COM from 28.27\% to 80.53\% and 81.67\%, respectively. These results demonstrate that incorporating style variance via CLUE substantially enhances the ability of generated images to capture the semantic characteristics of real samples, with particularly dramatic improvements for the COM category.

\begin{table}[h]
    \centering
    \caption{Recall scores (\%) between real and synthetic images}\label{tbl:recall}
    \centering
    \scriptsize
    \setlength{\tabcolsep}{4pt}
    \begin{tabular*}{\columnwidth}{@{\extracolsep{\fill}}lcccccccccccc@{}}
        \toprule
        \multirow{2}{*}{Model}
        & \multicolumn{4}{c}{Dataset 1 + Dataset 2}
        & \multicolumn{4}{c}{Dataset 1}
        & \multicolumn{4}{c}{Dataset 2} \\
        \cmidrule(lr){2-5}\cmidrule(lr){6-9}\cmidrule(lr){10-13}
        & Normal & OME & COM & All
        & Normal & OME & COM & All
        & Normal & OME & COM & All \\
        \midrule
        Vanilla            & 76.80 & 46.27 & 28.27 & 49.60 
                           & 81.42 & 50.17 & 30.83 & 51.83 
                           & 83.67 & 55.33 & 55.00 & 57.67 \\
        CLUE ($\sigma=0.0$)& 79.87 & 55.80 & 58.00 & 56.62 
                           & 84.08 & 59.92 & 62.67 & 59.19 
                           & 87.00 & 65.67 & 82.33 & 69.67 \\
        CLUE ($\sigma=0.1$)& 83.53 & 56.80 & 57.93 & 59.78 
                           & 87.25 & 61.17 & 62.75 & 62.53 
                           & 90.33 & 70.33 & 80.67 & 74.56 \\
        CLUE ($\sigma=0.3$)& \textbf{89.13} & 63.80 & 80.53 & 67.56 
                           & \textbf{91.17} & 68.92 & \textbf{83.83} & 69.97 
                           & 95.33 & 73.00 & 93.67 & 85.89 \\
        CLUE ($\sigma=0.5$)& 86.53 & \textbf{64.87} & \textbf{81.67} & \textbf{70.29} 
                           & 88.33 & \textbf{69.58} & 83.75 & \textbf{71.97} 
                           & \textbf{97.33} & \textbf{81.00} & \textbf{96.67} & \textbf{90.22} \\
        \bottomrule
    \end{tabular*}
\end{table}

\subsection{Classification Performance}
To evaluate the practical utility of CLUE generated images for downstream tasks, we conducted three sets of classification experiments. First, we assessed the impact of progressively replacing real training data with synthetic images. Second, we investigated the scaling behavior when training exclusively on synthetic data. Third, we explored the performance when combining synthetic data with real training data. In all experiments, Dataset 1B was used as validation set and Dataset 2 was used as the test set.

\subsubsection{Mixed Real–Synthetic Training}
Table~\ref{tbl:mixed_data_pm_all} summarizes the classification $F_1$ scores on Dataset 2 obtained by gradually replacing increasing proportions of real training images with synthetic ones, while keeping the validation set fixed. Trained exclusively on real data achieves a 93.81\% $F_1$ score. Substituting up to 20\% of the real images with synthetic samples yields consistent improvements across all model variants relative to this baseline. Although performance slowly declines beyond a 30\% replacement rate, it remains above 90\% even when 80\% of the training set is synthetic. Under a fully synthetic regime, the CLUE variant with $\sigma=0.5$ attains a 78.53\% $F_1$ score - an 8.28\% absolute gain over the vanilla's 70.25\%. Furthermore, the standard deviation for $\sigma=0.5$ under 1.08, demonstrating that CLUE generates more diverse and informative samples stably that better capture the underlying data distribution.

The Area Under the Receiver Operating Characteristic Curve (AUC) scores (Table~\ref{tbl:mixed_data_auc_all}) showed similar trends, with performance remaining above 98\% even at 80\% synthetic replacement, and CLUE with $\sigma=0.5$ maintaining the highest AUC of 98.50\%. At 100\% synthetic data, CLUE with $\sigma=0.5$ achieved 92.32\% AUC compared to the vanilla's 85.69\%, further confirming the framework's superior discriminative capability.

To further explore the potential of combining synthetic and real data, we conducted an extended experiment where we combined 100\% scale synthetic data (2,700 images) with the real Dataset 1A (2,700 images). This combined approach yielded exceptional results, with CLUE ($\sigma=0.5$) achieving an $F_1$ score of 94.76\%, surpassing all other variants and exceeding the performance of models trained on real data alone (93.81\%). The vanilla generation combined with real data achieved 94.44\%, while CLUE variants with $\sigma=0.0, 0.1$, and 0.3 achieved 94.23\%, 94.17\%, and 94.62\%, respectively. These results demonstrate that when high quality synthetic data is combined with real data, CLUE's style embeddings provide a consistent advantage, establishing its utility as a data augmentation technique in medical imaging applications.

%
\afterpage{%
  \clearpage
  \begin{table*}[h]
    \centering
    \caption{Classification $F_1$ score (\%) on Dataset 2 with different percentages of synthetic data replacement on train set \\ \footnotesize $^*$At 0\% synthetic data, classification model uses only real data.}
    \label{tbl:mixed_data_pm_all}
    {\footnotesize
    \begin{tabular*}{\textwidth}{@{\extracolsep{\fill}} l  l  l  l  l  l  @{} }
      \toprule
      Synthetic & Vanilla & $\sigma=0.0$ & $\sigma=0.1$ & $\sigma=0.3$ & $\sigma=0.5$ \\
      \midrule
      0 \%   & 93.81$^*$         &                            &                            &                              &   \\
      10 \%  & 94.11 ($\pm0.52$) & \textbf{94.44} ($\pm0.68$) & 94.09  ($\pm0.57$)         & 94.05  ($\pm0.57$)           & 94.00 ($\pm0.55$) \\
      20 \%  & 94.03 ($\pm0.61$) & 93.83 ($\pm0.52$)          & 93.98  ($\pm0.61$)         & \textbf{94.10} ($\pm0.57$)   & 94.08 ($\pm0.43$) \\
      30 \%  & 93.42 ($\pm0.59$) & 93.38 ($\pm0.52$)          & 93.13  ($\pm0.70$)         & 93.15  ($\pm0.97$)           & \textbf{93.72} ($\pm0.60$) \\
      40 \%  & 93.49 ($\pm0.74$) & 93.07 ($\pm0.91$)          & \textbf{93.69} ($\pm0.55$) & 93.68  ($\pm0.57$)           & 93.57 ($\pm0.52$) \\
      50 \%  & 92.88 ($\pm0.56$) & 92.77 ($\pm0.70$)          & 93.09  ($\pm0.51$)         & 92.81  ($\pm0.80$)           & \textbf{93.11} ($\pm0.44$) \\
      60 \%  & 92.04 ($\pm0.73$) & 92.58 ($\pm0.64$)          & 92.15  ($\pm0.68$)         & \textbf{92.73} ($\pm0.57$)   & 92.60 ($\pm0.60$) \\
      70 \%  & 91.80 ($\pm0.68$) & 91.92 ($\pm0.71$)          & 91.86  ($\pm0.74$)         & 92.00  ($\pm1.28$)           & \textbf{92.58} ($\pm0.53$) \\
      80 \%  & 90.67 ($\pm0.86$) & 90.51 ($\pm0.83$)          & 90.73  ($\pm0.93$)         & \textbf{91.11} ($\pm1.18$)   & 90.65 ($\pm0.93$) \\
      90 \%  & 88.08 ($\pm0.89$) & 87.04 ($\pm1.70$)          & 88.02  ($\pm1.34$)         & 90.03  ($\pm1.16$)           & \textbf{90.08} ($\pm0.72$) \\
      100 \% & 70.25 ($\pm4.19$) & 73.42 ($\pm2.63$)          & 70.99  ($\pm4.83$)         & 74.25  ($\pm4.68$)           & \textbf{78.53} ($\pm1.08$) \\
      \midrule
      100 \% + Dataset 1A & 94.44 ($\pm0.50$) & 94.23 ($\pm0.52$) & 94.17 ($\pm0.67$) & 94.62 ($\pm0.43$) & \textbf{94.76} ($\pm0.50$) \\
      \bottomrule
    \end{tabular*}}
  \end{table*}

  \begin{table*}[h]
    \centering
    \caption{Classification AUC score (\%) on Dataset 2 with different percentages of synthetic data replacement on train set \\ \footnotesize $^*$At 0\% synthetic data, classification model uses only real data.}
    \label{tbl:mixed_data_auc_all}
    {\footnotesize
    \begin{tabular*}{\textwidth}{@{\extracolsep{\fill}} l  l  l  l  l  l  @{} }
      \toprule
      Synthetic ratio & Vanilla & $\sigma=0.0$ & $\sigma=0.1$ & $\sigma=0.3$ & $\sigma=0.5$ \\
      \midrule
      0 \%   & 99.06$^*$         &                            &                   &                            & \\
      10 \%  & 98.90 ($\pm0.23$) & \textbf{98.94} ($\pm0.11$) & 98.87 ($\pm0.24$) & 98.77 ($\pm0.26$)          & 98.78 ($\pm0.13$) \\
      20 \%  & 98.79 ($\pm0.21$) & \textbf{98.92} ($\pm0.18$) & 98.90 ($\pm0.20$) & 98.85 ($\pm0.21$)          & 98.91 ($\pm0.21$) \\
      30 \%  & 98.74 ($\pm0.17$) & 98.75 ($\pm0.14$)          & 98.72 ($\pm0.19$) & 98.83 ($\pm0.29$)          & \textbf{98.87} ($\pm0.16$) \\
      40 \%  & 98.77 ($\pm0.20$) & 98.78 ($\pm0.18$)          & 98.77 ($\pm0.21$) & 98.79 ($\pm0.23$)          & \textbf{98.87} ($\pm0.14$) \\
      50 \%  & 98.51 ($\pm0.20$) & 98.49 ($\pm0.22$)          & 98.63 ($\pm0.24$) & 98.59 ($\pm0.27$)          & \textbf{98.66} ($\pm0.23$) \\
      60 \%  & 98.40 ($\pm0.23$) & 98.46 ($\pm0.24$)          & 98.35 ($\pm0.14$) & \textbf{98.56} ($\pm0.13$) & 98.45 ($\pm0.20$) \\
      70 \%  & 98.16 ($\pm0.42$) & 98.21 ($\pm0.24$)          & 98.23 ($\pm0.32$) & 98.22 ($\pm0.44$)          & \textbf{98.50} ($\pm0.25$) \\
      80 \%  & 97.87 ($\pm0.38$) & 97.75 ($\pm0.46$)          & 98.05 ($\pm0.38$) & 98.10 ($\pm0.47$)          & \textbf{98.11} ($\pm0.38$) \\
      90 \%  & 96.73 ($\pm0.42$) & 96.19 ($\pm0.78$)          & 96.86 ($\pm0.41$) & 97.63 ($\pm0.44$)          & \textbf{97.64} ($\pm0.41$) \\
      100 \% & 85.69 ($\pm2.20$) & 88.36 ($\pm1.71$)          & 86.50 ($\pm3.46$) & 89.38 ($\pm2.91$)          & \textbf{92.32} ($\pm1.01$) \\
      \midrule
      100 \% + Dataset 1A & 98.92 ($\pm0.16$) & 98.97 ($\pm0.13$) & 98.93 ($\pm0.18$) & 99.07 ($\pm0.17$) & \textbf{99.09} ($\pm0.14$) \\
      \bottomrule
    \end{tabular*}}
  \end{table*}

  \begin{figure*}[h]
    \centering
    \includegraphics[width=1\textwidth]{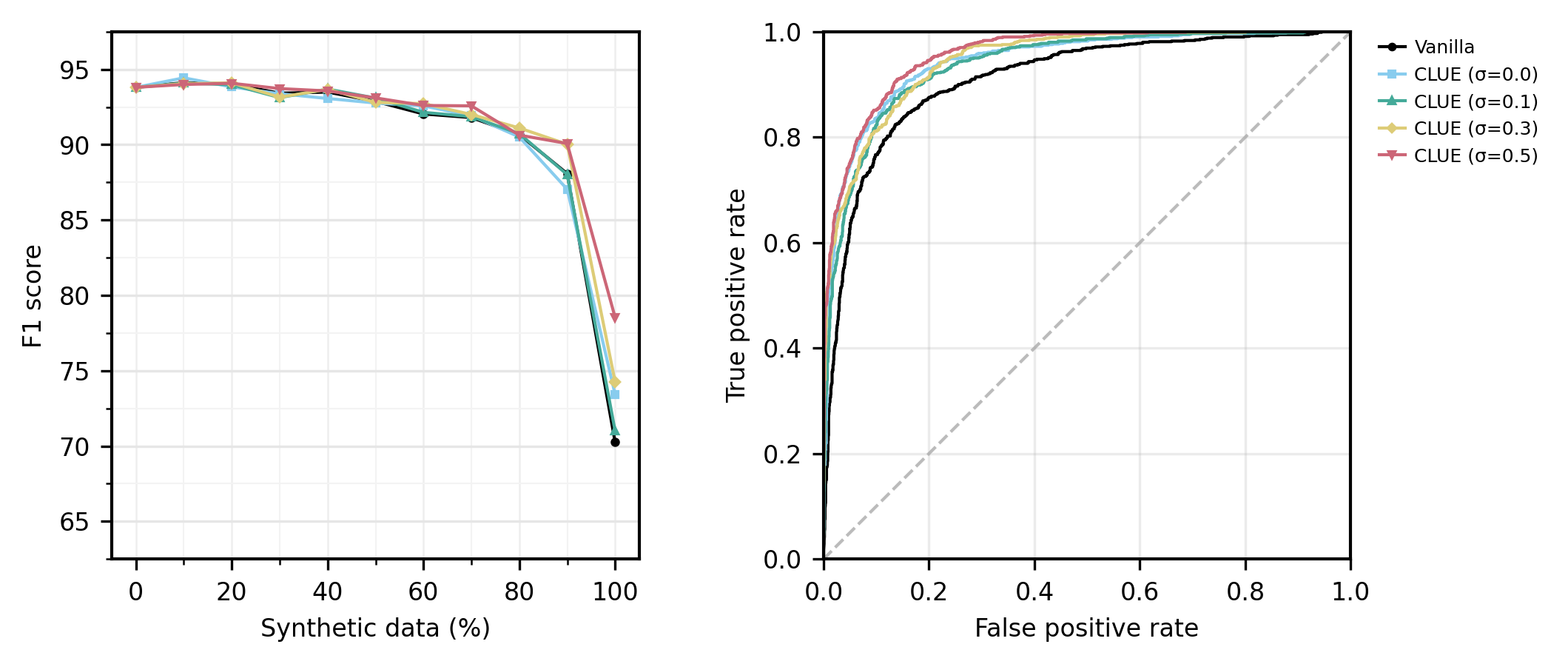}
    \caption{Dataset 2 (Left) $F_1$ score with mixed real-synthetic, (Right) ROC curves at 100\% synthetic}
    \label{fig:mix}
  \end{figure*}
}

\afterpage{%
  \clearpage
  \begin{table*}[h]
    \centering
    \caption{Classification $F_1$ score (\%) on AI-HUB dataset with different percentages of synthetic data replacement on train set \\ \footnotesize $^*$At 0\% synthetic data, classification model uses only real data.}
    \label{tbl:mixed_data_pm_all_aihub}
    {\footnotesize
    \begin{tabular*}{\textwidth}{@{\extracolsep{\fill}} l  l  l  l  l  l  @{} }
      \toprule
      Synthetic & Vanilla & $\sigma=0.0$ & $\sigma=0.1$ & $\sigma=0.3$ & $\sigma=0.5$ \\
      \midrule
      0 \% & \textbf{83.76 ($\pm0.00$)$^*$} &  &  &  &  \\
      10 \% & 83.94 ($\pm1.22$) & 83.45 ($\pm1.60$) & 83.66 ($\pm1.56$) & 83.83 ($\pm1.04$) & \textbf{84.81 }($\pm0.80$) \\
      20 \% & 83.97 ($\pm1.02$) & 84.11 ($\pm1.74$) & 83.53 ($\pm1.34$) & 84.56 ($\pm1.53$) & \textbf{85.30 }($\pm1.03$) \\
      30 \% & 83.33 ($\pm1.46$) & 83.87 ($\pm0.89$) & 83.70 ($\pm1.11$) & 84.20 ($\pm1.18$) & \textbf{85.03 }($\pm0.99$) \\
      40 \% & 83.20 ($\pm1.60$) & 83.30 ($\pm1.30$) & 83.49 ($\pm1.33$) & 83.94 ($\pm1.19$) & \textbf{85.29 }($\pm0.35$) \\
      50 \% & 82.37 ($\pm1.54$) & 81.06 ($\pm2.05$) & 81.63 ($\pm1.12$) & 82.99 ($\pm0.85$) & \textbf{84.10 }($\pm1.19$) \\
      60 \% & 81.18 ($\pm1.40$) & 80.59 ($\pm1.84$) & 81.47 ($\pm1.76$) & 82.70 ($\pm1.52$) & \textbf{83.49 }($\pm0.80$) \\
      70 \% & 80.96 ($\pm0.97$) & 79.11 ($\pm2.09$) & 80.39 ($\pm1.74$) & 81.85 ($\pm1.16$) & \textbf{82.96 }($\pm1.05$) \\
      80 \% & 77.75 ($\pm2.52$) & 76.88 ($\pm2.37$) & 78.31 ($\pm2.10$) & 80.61 ($\pm1.13$) & \textbf{81.53 }($\pm1.29$) \\
      90 \% & 75.73 ($\pm2.42$) & 74.55 ($\pm1.80$) & 76.21 ($\pm1.23$) & 78.81 ($\pm2.63$) & \textbf{80.94 }($\pm0.78$) \\
      100 \% & 54.26 ($\pm5.28$) & 59.48 ($\pm4.87$) & 59.87 ($\pm7.02$) & 66.61 ($\pm4.60$) & \textbf{72.37 }($\pm2.80$) \\
      \midrule
      100 \% + Dataset 1A & 84.67 ($\pm1.17$) & 83.63 ($\pm1.06$) & 83.48 ($\pm1.28$) & 85.23 ($\pm0.58$) & \textbf{85.78} ($\pm1.42$) \\
      \bottomrule
    \end{tabular*}}
  \end{table*}

  \begin{table*}[h]
    \centering
    \caption{Classification AUC score (\%) on AI-HUB dataset with different percentages of synthetic data replacement on train set \\ \footnotesize $^*$At 0\% synthetic data, classification model uses only real data.}
    \label{tbl:mixed_data_auc_all_aihub}
    {\footnotesize
    \begin{tabular*}{\textwidth}{@{\extracolsep{\fill}} l  l  l  l  l  l  @{} }
      \toprule
      Synthetic ratio & Vanilla & $\sigma=0.0$ & $\sigma=0.1$ & $\sigma=0.3$ & $\sigma=0.5$ \\
      \midrule
      0 \% & 94.86$^*$ \\
      10 \% & 94.67 ($\pm0.44$) & 94.63 ($\pm0.77$) & 94.54 ($\pm0.67$) & 94.76 ($\pm0.49$) & \textbf{94.92} ($\pm0.40$) \\
      20 \% & 94.49 ($\pm0.60$) & 94.65 ($\pm0.80$) & 94.47 ($\pm0.63$) & 94.81 ($\pm0.69$) & \textbf{95.24} ($\pm0.50$) \\
      30 \% & 94.24 ($\pm0.94$) & 94.65 ($\pm0.43$) & 94.50 ($\pm0.61$) & 94.71 ($\pm0.69$) & \textbf{95.19} ($\pm0.50$) \\
      40 \% & 94.20 ($\pm0.97$) & 94.29 ($\pm0.59$) & 94.36 ($\pm0.45$) & 94.62 ($\pm0.61$) & \textbf{95.16} ($\pm0.16$) \\
      50 \% & 93.85 ($\pm0.82$) & 93.35 ($\pm0.92$) & 93.57 ($\pm0.34$) & 94.24 ($\pm0.54$) & \textbf{94.70} ($\pm0.64$) \\
      60 \% & 93.23 ($\pm0.75$) & 92.66 ($\pm1.03$) & 93.42 ($\pm0.78$) & 94.05 ($\pm0.63$) & \textbf{94.44} ($\pm0.42$) \\
      70 \% & 93.06 ($\pm0.65$) & 92.39 ($\pm0.89$) & 92.80 ($\pm0.91$) & 93.54 ($\pm0.72$) & \textbf{94.17} ($\pm0.53$) \\
      80 \% & 91.18 ($\pm1.46$) & 90.94 ($\pm1.55$) & 91.73 ($\pm1.04$) & 92.83 ($\pm0.74$) & \textbf{93.43} ($\pm0.71$) \\
      90 \% & 89.62 ($\pm1.76$) & 88.70 ($\pm1.54$) & 90.28 ($\pm0.71$) & 91.83 ($\pm1.39$) & \textbf{93.17} ($\pm0.63$) \\
      100 \% & 74.90 ($\pm2.79$) & 77.91 ($\pm4.04$) & 78.04 ($\pm4.91$) & 83.14 ($\pm3.40$) & \textbf{88.02} ($\pm2.18$) \\
      \midrule
      100 \% + Dataset 1A & 94.96 ($\pm0.60$) & 94.66 ($\pm0.31$) & 94.54 ($\pm0.72$) & 95.21 ($\pm0.35$) & \textbf{95.43} ($\pm0.63$) \\
      \bottomrule
    \end{tabular*}}
  \end{table*}

  \begin{figure*}[h]
    \centering
    \includegraphics[width=1\textwidth]{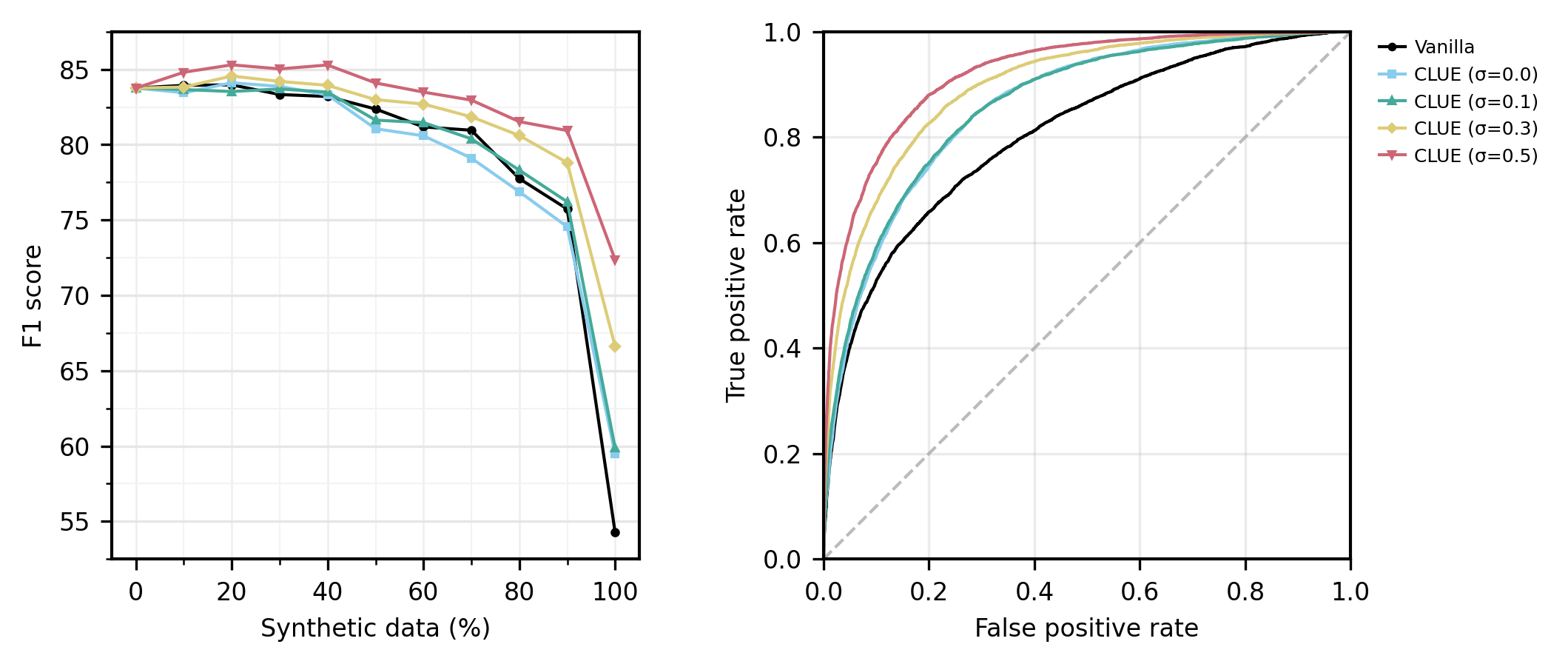}
    \caption{AI-HUB dataset (Left) $F_1$ score with mixed real-synthetic, (Right) ROC curves at 100\% synthetic}
    \label{fig:mix_aihub}
  \end{figure*}

}

Table~\ref{tbl:mixed_data_pm_all_aihub} presents the evaluation results on the external AI-HUB dataset using models trained on Dataset 1A with varying proportions of synthetic data. The baseline model trained solely on Dataset 1A achieves an 83.76\% $F_1$ score on the AI-HUB test set, reflecting the challenges of cross-institution generalization. Remarkably, CLUE with $\sigma=0.5$ achieves $F_1$ scores ranging from 84.81\% to 85.30\% at 10-40\% synthetic data replacement, surpassing the Dataset 1A-only baseline by up to 1.54\% points. This suggests that synthetic samples generated by CLUE do not simply replicate the source data but introduce beneficial variations that enhance the model's cross-institution generalization capabilities. While performance degradation is observed beyond 50\% synthetic replacement, CLUE ($\sigma=0.5$) maintains superior performance compared to other variants. Even at 80\% synthetic data, it retains an 81.53\% $F_1$ score, and when trained entirely on synthetic data, it achieves 72.37\% - 18.11\% higher than the vanilla's 54.26\%. This demonstrates that CLUE generated synthetic data provides better generalization to external test sets.

The AUC scores (Table~\ref{tbl:mixed_data_auc_all_aihub}) show more robust performance. CLUE with $\sigma=0.5$ maintains 93.17\% AUC even at 90\% synthetic replacement and achieves 88.02\% at 100\% synthetic data compared to the vanilla's 74.90\%. The consistently lower standard deviations across all metrics for CLUE indicate more stable and reliable performance on external datasets. The final row showing 100\% synthetic data + Dataset 1A results is particularly noteworthy. CLUE with $\sigma=0.5$ achieves 85.78\% $F_1$ score and 95.43\% AUC, outperforming the Dataset 1A-only baseline. This suggests that synthetic data can complement real data to enhance the model's cross-institution generalization capabilities.

These results demonstrate that the CLUE framework does not merely generate samples overfitted to the source domain, but produces diverse and representative synthetic samples that generalize well to external datasets with different acquisition protocols and patient populations. The consistent improvements on the AI-HUB dataset, despite being trained on a different dataset, validate CLUE's effectiveness in creating synthetic data that captures generalizable features rather than dataset-specific artifacts.

\subsubsection{Synthetic Only Training at Scale}
To investigate whether increased data volume could compensate for the quality gap between real and synthetic images, we trained classifiers using only synthetic data at various scales (Tables~\ref{tbl:scaling_extended}). The original training set size of 2,700 images (900 per class) was scaled from 200\% up to 1000\%, generating up to 27,000 synthetic images.

\afterpage{%
    \clearpage
    \begin{table*}[h]
        \centering
        \caption{$F_1$ score (\%) on Dataset 2, only using synthetic data for classification train set} 
        \label{tbl:scaling_extended}    
        \begin{tabular*}{\columnwidth}{@{\extracolsep{\fill}} l  l  l  l  l  l  @{} }
            \toprule
            Dataset Scale & Vanilla & $\sigma=0.0$ & $\sigma=0.1$ & $\sigma=0.3$ & $\sigma=0.5$ \\
            \midrule
            100 \%  & 70.25 ($\pm4.19$)  & 73.42 ($\pm2.63$)  & 70.99 ($\pm4.83$)  & 74.25 ($\pm4.68$) & \textbf{78.53} ($\pm1.08$) \\
            200 \%  & 70.60 ($\pm4.02$)  & 74.32 ($\pm4.35$)  & 72.09 ($\pm5.43$)  & 75.37 ($\pm5.02$) & \textbf{80.43} ($\pm3.58$) \\
            500 \%  & 74.68 ($\pm3.17$)  & 76.22 ($\pm3.81$)  & 70.41 ($\pm8.22$) & 80.32 ($\pm2.96$) & \textbf{81.82} ($\pm1.92$) \\
            1000 \% & 73.83 ($\pm5.20$) & 73.65 ($\pm5.20$) & 73.92 ($\pm7.54$) & 80.73 ($\pm1.72$) & \textbf{83.21} ($\pm2.16$) \\
            \midrule
            1000 \% + Dataset 1A & 92.15 ($\pm0.79$) & 92.08 ($\pm0.52$) & 92.32 ($\pm0.72$) & 92.63 ($\pm0.80$) & \textbf{93.41} ($\pm0.65$) \\
            \bottomrule
        \end{tabular*}
    \end{table*}
    \begin{table*}[h]
        \centering
        \caption{AUC score (\%) on Dataset 2, only using synthetic data for classification train set}
        \label{tbl:scaling_auc_extended}
        \begin{tabular*}{\columnwidth}{@{\extracolsep{\fill}} l  l  l  l  l  l  @{} }
            \toprule
            Dataset Scale & Vanilla & $\sigma=0.0$ & $\sigma=0.1$ & $\sigma=0.3$ & $\sigma=0.5$ \\
            \midrule
            100 \%  & 85.69 ($\pm2.20$) & 88.36 ($\pm1.71$) & 86.50 ($\pm3.46$) & 89.38 ($\pm2.91$) & \textbf{92.32} ($\pm1.01$) \\
            200 \%  & 85.98 ($\pm3.34$) & 89.25 ($\pm2.37$) & 87.81 ($\pm2.90$) & 90.11 ($\pm2.83$) & \textbf{93.57} ($\pm1.51$) \\
            500 \%  & 88.76 ($\pm2.21$) & 89.73 ($\pm2.76$) & 86.59 ($\pm5.49$) & 93.18 ($\pm1.65$) & \textbf{94.13} ($\pm0.93$) \\
            1000 \% & 87.62 ($\pm3.99$) & 87.78 ($\pm3.68$) & 88.91 ($\pm5.12$) & 93.02 ($\pm0.95$) & \textbf{94.78} ($\pm1.32$) \\
            \midrule
            1000 \% + Dataset 1A & 98.28 ($\pm0.32$) & 98.16 ($\pm0.44$) & 98.47 ($\pm0.28$) & 98.64 ($\pm0.33$) & \textbf{98.70} ($\pm0.28$) \\
            \bottomrule
        \end{tabular*}
    \end{table*}
    \begin{figure*}[h]
        \centering
        \includegraphics[width=1\columnwidth]{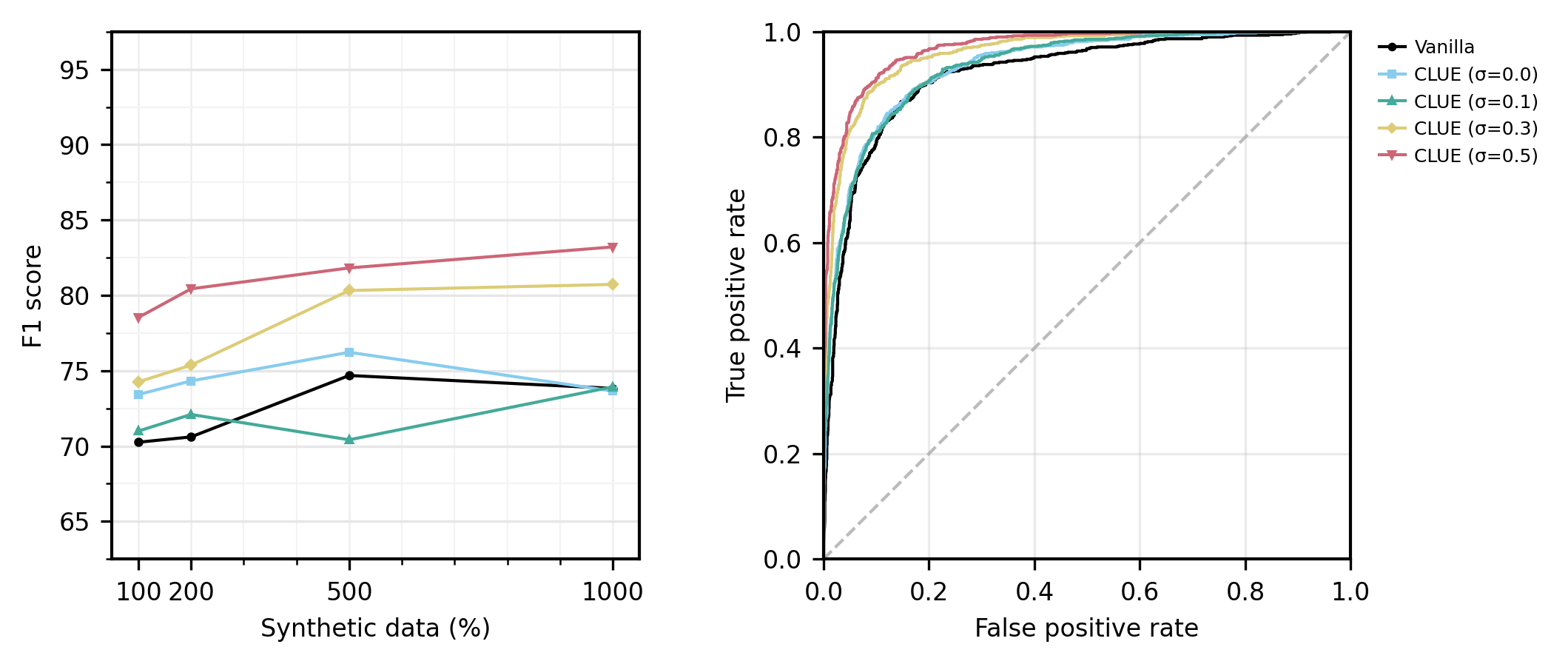}
        \caption{Dataset 2 (Left) $F_1$ score with synthetic only at various scales, (Right) ROC curves at 1000\% synthetic}\label{fig:scale}
    \end{figure*}

}

The scaling behavior reveals distinct patterns across model variants. CLUE with $\sigma=0.5$ continued to benefit from additional data, improving from 78.53\% at 100\% scale to 83.21\% at 1000\% scale. This 4.68\% improvement through scaling suggests that CLUE's style embeddings enable generation of meaningfully diverse samples that provide additional training signal even at large scales. The standard deviation for $\sigma=0.5$ remains stable at 3.58, while others exceed 5.00. CLUE with $\sigma=0.3$ also shows performance improvements with scale. In contrast, the vanilla and CLUE with lower sigma fail to demonstrate consistent scaling trends. This indicates that insufficient diversity in synthetic data generation limits the ability to establish a clear relationship between training volume and performance improvements. CLUE with high sigma values can address this limitation by maintaining sufficient style diversity throughout the scaling process.

The AUC scores (Table~\ref{tbl:scaling_auc_extended}) further validated these findings, with CLUE $\sigma=0.5$ achieving 94.78\% at 1000\% scale compared to the vanilla's 87.62\%, demonstrating superior discriminative capability. Figure~\ref{fig:mix} visualizes the $F_1$ score trends for both mixed training and synthetic only scenarios, while Figure~\ref{fig:scale} presents the corresponding scaling performance. Additionally, when combining 1000\% scale synthetic data (27,000 images) with the real Dataset 1A (2,700 images), CLUE with $\sigma=0.5$ achieved an $F_1$ score of 93.41\% and an AUC of 98.70\%, further validating the framework's capability for large scale synthetic data generation and its complementary value when combined with real data.

\afterpage{%
    \clearpage
    \begin{table*}[h]
        \centering
        \caption{$F_1$ score (\%) on AI-HUB dataset, only using synthetic data for classification train set} 
        \label{tbl:scaling_extended_aihub}    
        \begin{tabular*}{\columnwidth}{@{\extracolsep{\fill}} l  l  l  l  l  l  @{} }
            \toprule
            Dataset Scale & Vanilla & $\sigma=0.0$ & $\sigma=0.1$ & $\sigma=0.3$ & $\sigma=0.5$ \\
            \midrule
            100 \% & 54.26 ($\pm5.28$) & 59.48 ($\pm4.87$) & 59.87 ($\pm7.02$) & 66.61 ($\pm4.60$) & \textbf{72.37 }($\pm2.80$) \\
            200 \% & 54.60 ($\pm3.48$) & 58.81 ($\pm5.08$) & 58.71 ($\pm6.51$) & 67.75 ($\pm6.28$) & \textbf{73.69 }($\pm3.13$) \\
            500 \% & 62.95 ($\pm3.94$) & 61.09 ($\pm5.30$) & 54.74 ($\pm7.32$) & 73.22 ($\pm2.76$) & \textbf{75.64 }($\pm2.07$) \\
            1000 \% & 60.61 ($\pm5.39$) & 59.38 ($\pm5.92$) & 63.18 ($\pm6.05$) & 73.29 ($\pm1.62$) & \textbf{76.77 }($\pm1.83$) \\
            \midrule
            1000 \% + Dataset 1A & 79.67 ($\pm2.27$) & 78.44 ($\pm1.88$) & 77.83 ($\pm2.11$) & 81.9 ($\pm1.13$) & \textbf{84.02} ($\pm0.99$) \\
            \bottomrule
        \end{tabular*}
    \end{table*}
    \begin{table*}[h]
        \centering
        \caption{AUC score (\%) on AI-HUB dataset, only using synthetic data for classification train set}
        \label{tbl:scaling_auc_extended_aihub}
        \begin{tabular*}{\columnwidth}{@{\extracolsep{\fill}} l  l  l  l  l  l  @{} }
            \toprule
            Dataset Scale & Vanilla & $\sigma=0.0$ & $\sigma=0.1$ & $\sigma=0.3$ & $\sigma=0.5$ \\
            \midrule
            100 \% & 74.90 ($\pm2.79$) & 77.91 ($\pm4.04$) & 78.04 ($\pm4.91$) & 83.14 ($\pm3.40$) & \textbf{88.02} ($\pm2.18$) \\
            200 \% & 74.59 ($\pm3.44$) & 76.47 ($\pm4.60$) & 77.69 ($\pm3.77$) & 84.25 ($\pm4.14$) & \textbf{89.23} ($\pm1.94$) \\
            500 \% & 80.76 ($\pm2.96$) & 78.88 ($\pm4.25$) & 75.13 ($\pm5.36$) & 88.46 ($\pm2.00$) & \textbf{90.06} ($\pm1.49$) \\
            1000 \% & 77.86 ($\pm4.82$) & 76.33 ($\pm4.59$) & 80.12 ($\pm4.43$) & 88.08 ($\pm1.51$) & \textbf{90.85} ($\pm1.57$) \\
            \midrule
            1000 \% + Dataset 1A & 92.10 ($\pm1.44$) & 91.54 ($\pm1.41$) & 91.52 ($\pm1.31$) & 93.81 ($\pm0.59$) & \textbf{94.58} ($\pm0.79$) \\
            \bottomrule
        \end{tabular*}
    \end{table*}

    \begin{figure*}[h]
        \centering
        \includegraphics[width=1\columnwidth]{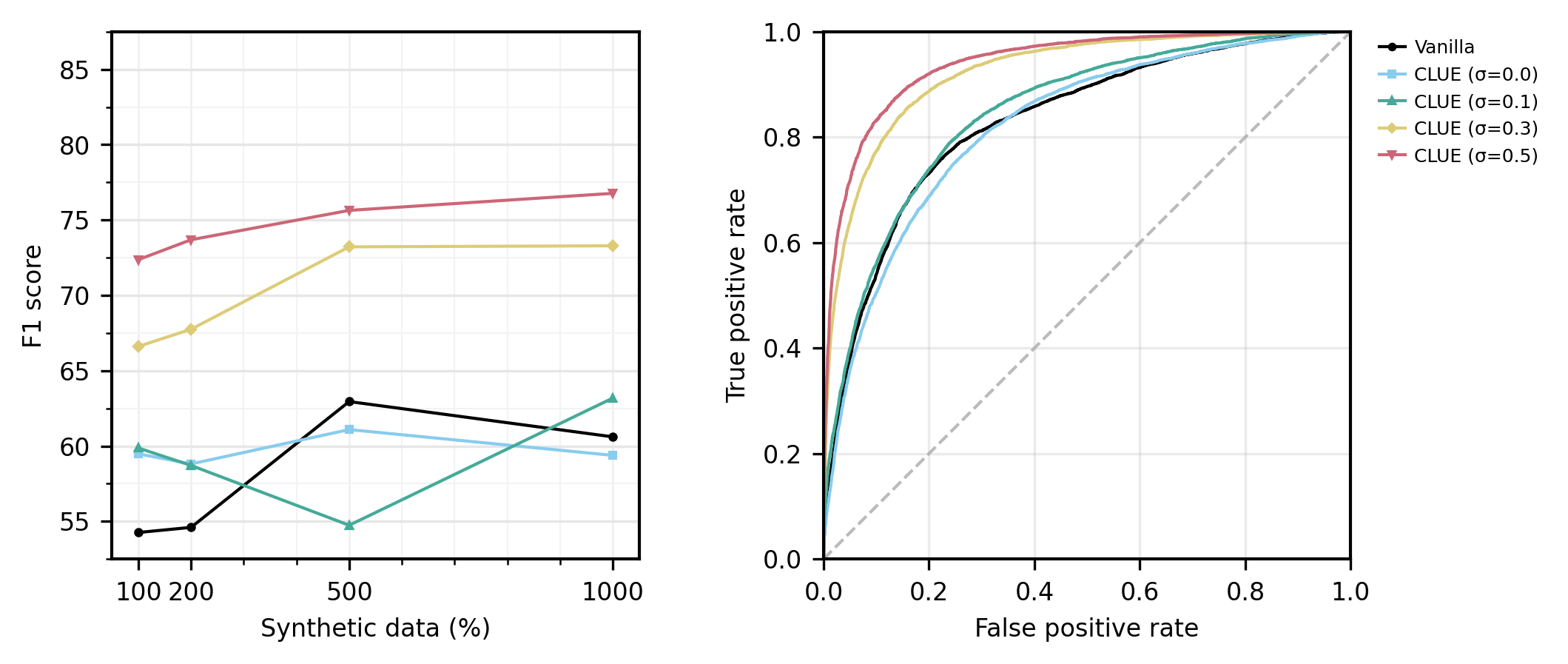}
        \caption{AI-HUB dataset (Left) $F_1$ score with synthetic only at various scales, (Right) ROC curves at 1000\% synthetic}\label{fig:scale_aihub}
    \end{figure*}
    \clearpage

}

These classification results validate that CLUE can produce dataset that preserves diagnostic features essential for clinical decision-making. The framework's ability to approach real-data performance levels when provided with sufficient synthetic samples positions it as a practical solution for data augmentation in medical imaging contexts.

To evaluate the cross-institution scalability of synthetic data generation, we tested models trained exclusively on synthetic data at various scales on the external AI-HUB dataset (Table~\ref{tbl:scaling_extended_aihub}). This challenging evaluation scenario assesses whether increased synthetic data volume can improve generalization to unseen external datasets.

At the baseline scale (100\%), CLUE with $\sigma=0.5$ achieves 72.37\% $F_1$ score, substantially outperforming the vanilla model's 54.26\% by 18.11\%. This performance with increased data volume: at 1000\% scale (27,000 synthetic images), CLUE with $\sigma=0.5$ reaches 76.77\%, demonstrating a 4.40\% improvement through scaling alone. In contrast, the vanilla model shows 60.61\% at 1000\% scale, suggesting that simply increasing quantity without diversity provides limited benefit for cross-institution generalization. The scaling patterns reveal important insights about synthetic data quality. While CLUE with $\sigma=0.3$ also shows consistent improvement 66.61\% to 73.29\%, the $\sigma=0.5$ variant maintains superior performance across all scales with lower standard deviations, indicating more stable and reliable cross-institution transfer.

AUC scores (Table~\ref{tbl:scaling_auc_extended_aihub}) further validate these findings. CLUE with $\sigma=0.5$ improves from 88.02\% at 100\% scale to 90.85\% at 1000\% scale, while maintaining consistently lower variance than other variants. The vanilla's AUC plateaus around 77.86\%, highlighting the limitations of style-agnostic generation for external validation. Notably, when combining 1000\% scale synthetic data with the original Dataset 1A (bottom row), CLUE with $\sigma=0.5$ achieves 84.02\% $F_1$ score and 94.58\% AUC. 

The consistent improvements on the AI-HUB dataset despite collected from the different institution than Dataset 1A validate that CLUE captures generalizable diagnostic features rather than dataset-specific artifacts. The framework's ability to improve cross-institution performance through scaling positions it as a practical solution for scenarios where acquiring domain-specific training data is challenging or expensive.

\section{Conclusion}
We introduced CLUE, Controllable Latent space of Unprompted Embeddings, a novel framework that addresses the fundamental challenge of controlling unprompted visual attributes in medical image generation. By augmenting Stable Diffusion with a Style Encoder that learns disentangled visual representations, CLUE enables fine-grained control over image characteristics that cannot be adequately expressed through text alone—a critical capability for medical imaging where subtle visual variations carry diagnostic significance.

Our comprehensive evaluation on otitis media datasets demonstrates CLUE's effectiveness across multiple dimensions. The framework achieved substantial improvements in generation quality, reducing FID from 46.81 to 9.30 and increasing recall from 49.60\% to 70.29\%. These metrics validate that CLUE generates images that are both visually realistic and semantically diverse, better capturing the natural variation present in clinical data. The progressive improvements with increasing style variance $\sigma$ confirm that controlled diversity is essential for comprehensive representation of medical image distributions.

The practical utility of CLUE-generated images is evidenced by strong classification performance across various training scenarios. When replacing portions of real training data with synthetic images, CLUE maintained robust performance even at 80\% synthetic replacement, achieving 90.65\% $F_1$ score with $\sigma=0.5$. Models trained exclusively on synthetic data at 1000\% scale achieved 83.21\% $F_1$ score. Most notably, combining 100\% scale synthetic data with equal amounts of real Dataset 1A yielded exceptional results: CLUE with $\sigma=0.5$ achieved 94.76\% $F_1$ score, surpassing both the real-data-only baseline of 93.81\% and vanilla generation combined with real data at 94.44\%. This demonstrates CLUE's superior effectiveness as a data augmentation strategy.

Particularly noteworthy is CLUE's cross-institution generalization capability. On the external AI-HUB dataset, models trained exclusively on CLUE-generated synthetic data achieved 76.77\% $F_1$ score at 1000\% scale, compared to vanilla's 60.61\%. When combined with real Dataset 1A, CLUE with $\sigma=0.5$ reached 85.78\% $F_1$ score, exceeding the 83.76\% achieved by models trained on real Dataset 1A alone. This robust transfer to external datasets with different acquisition protocols validates that CLUE learns generalizable representations rather than dataset-specific artifacts, while the synergistic effect of combining synthetic and real data demonstrates its practical value for improving model robustness across institutions.

While our evaluation focused on otitis media imaging, the underlying principle of learning controllable unprompted embeddings has broader implications for medical image synthesis. The framework's ability to generate diverse yet diagnostically consistent images without requiring additional conditioning inputs addresses key challenges in medical AI development, particularly in specialized domains where diverse training data and detailed annotations are scarce.

\subsection{Limitation}
Despite these promising results, our study has several limitations. First, our evaluation was confined to otitis media imaging with endoscopic tympanic membrane photographs. Second, our experiments were conducted on a relatively balanced dataset with sufficient samples per class. The classification model already demonstrated high performance on this balanced dataset. Third, the style embedding space, while effective for capturing visual variations, lacks explicit clinical interpretability. Finally, our evaluation relied primarily on automated metrics and classification performance; clinical validation through expert assessment would provide additional insights into practical applicability.

\subsection{Future Work}
While CLUE demonstrates strong potential, several directions remain open for future research. First, CLUE can be extended to a wider range of medical imaging modalities to evaluate its generalizability beyond otitis media. Second, its effectiveness in class imbalanced settings should be assessed, particularly to determine whether synthetic data generated by CLUE can mitigate performance degradation and enhance $F_1$ scores. Third, re-implementing CLUE using more advanced diffusion architectures—such as Stable Diffusion 3, which integrates multimodal diffusion transformers (MM-DiT)—may further improve both generation quality and control. These extensions would broaden the applicability and robustness of CLUE in diverse clinical and technical contexts.






\clearpage

\appendix
\section{Examples of the Real Dataset}\label{sec:appendixA}
    \begin{figure}[ht]
        \centering
        \includegraphics[width=0.8\columnwidth]{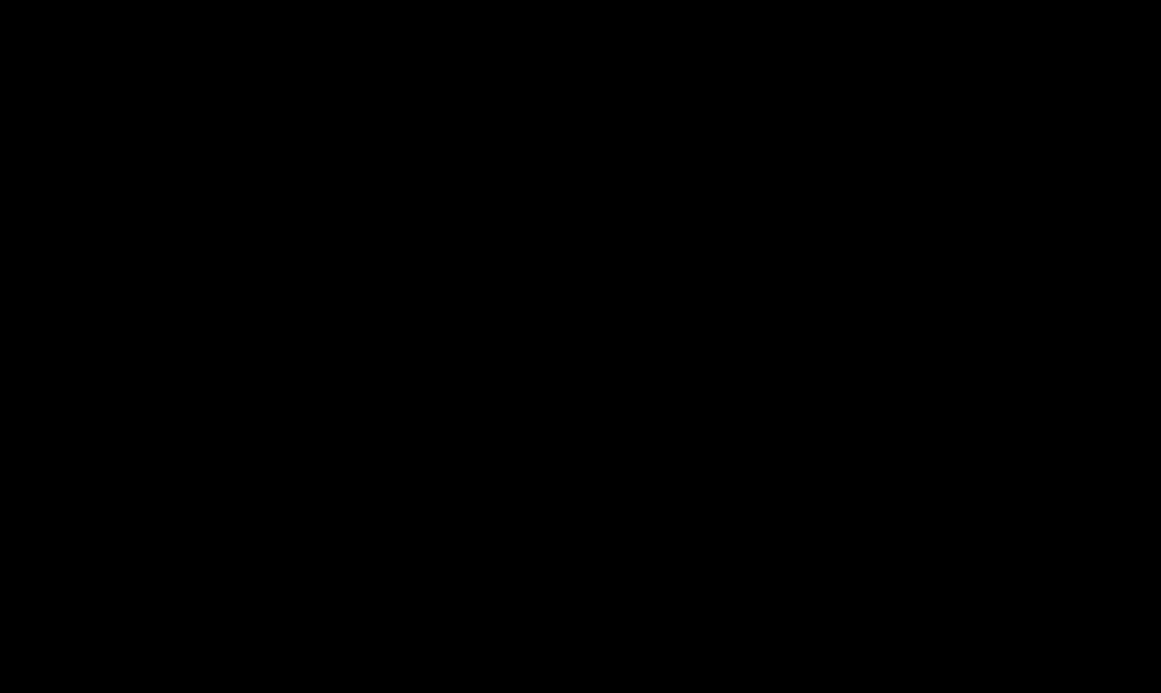}
        \caption{Samples of the real dataset.
        The rows correspond to: (1) Normal, (2) OME, (3) COM.}        \label{dataex}
    \end{figure}

\section{Learning curves of Generative Models}\label{sec:appendixB}
    \begin{figure}[ht]
        \centering
        \includegraphics[width=1\columnwidth]{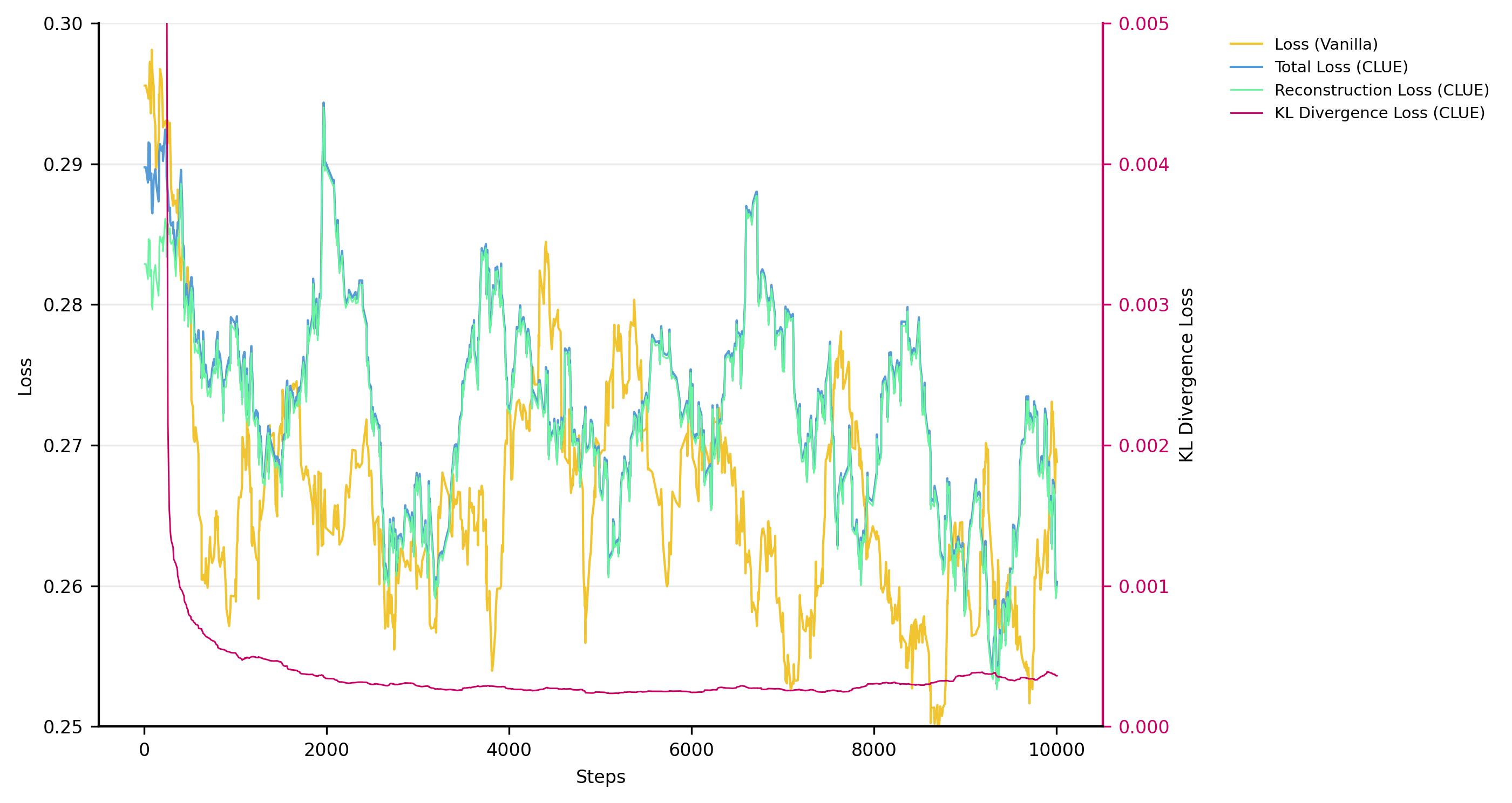}
        \caption{Learning curve. Total loss shows the summation of the CLUE's denoising loss and KL divergence loss.}
        \label{learncurv}
    \end{figure}

\clearpage

\section{Examples of Synthetic Datasets}\label{sec:appendixC}
\subsection{Vanilla}
\begin{figure}[ht]
  \centering
  \includegraphics[width=0.8\columnwidth]{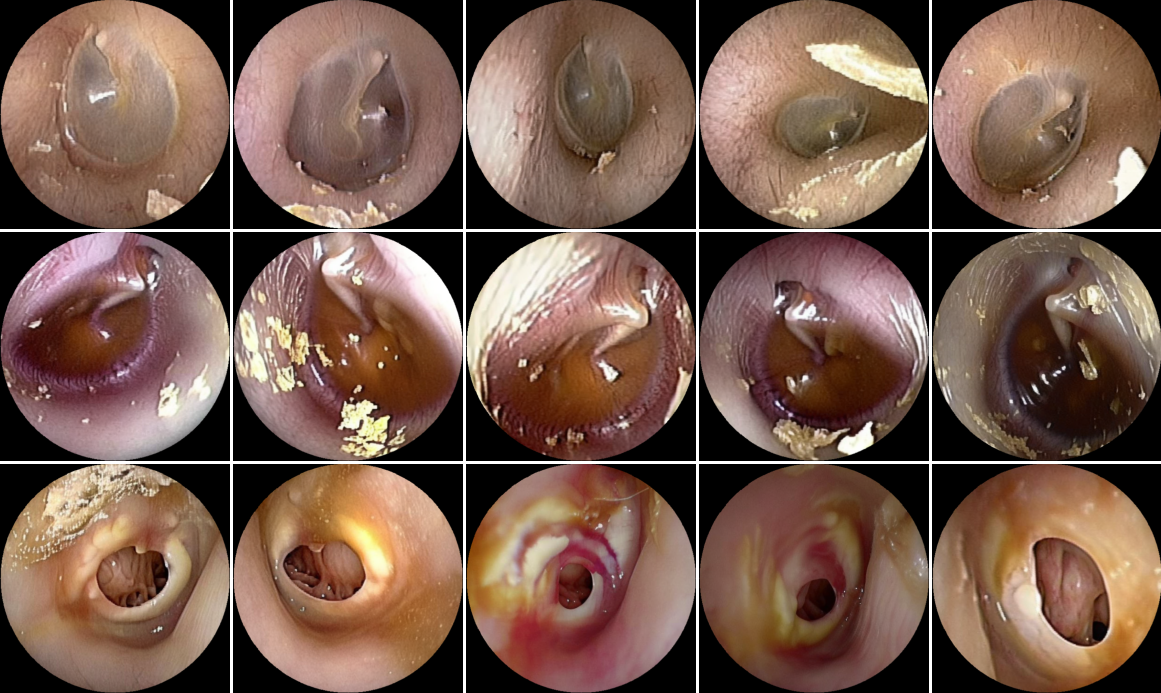}
  \caption{Samples of the vanilla synthetic dataset.
  The rows correspond to: (1) Normal, (2) OME, (3) COM.}
  \label{fig:vanilla_samples}
\end{figure}

\subsection{CLUE ($\sigma=0.0$)}
\begin{figure}[ht]
  \centering
  \includegraphics[width=0.8\columnwidth]{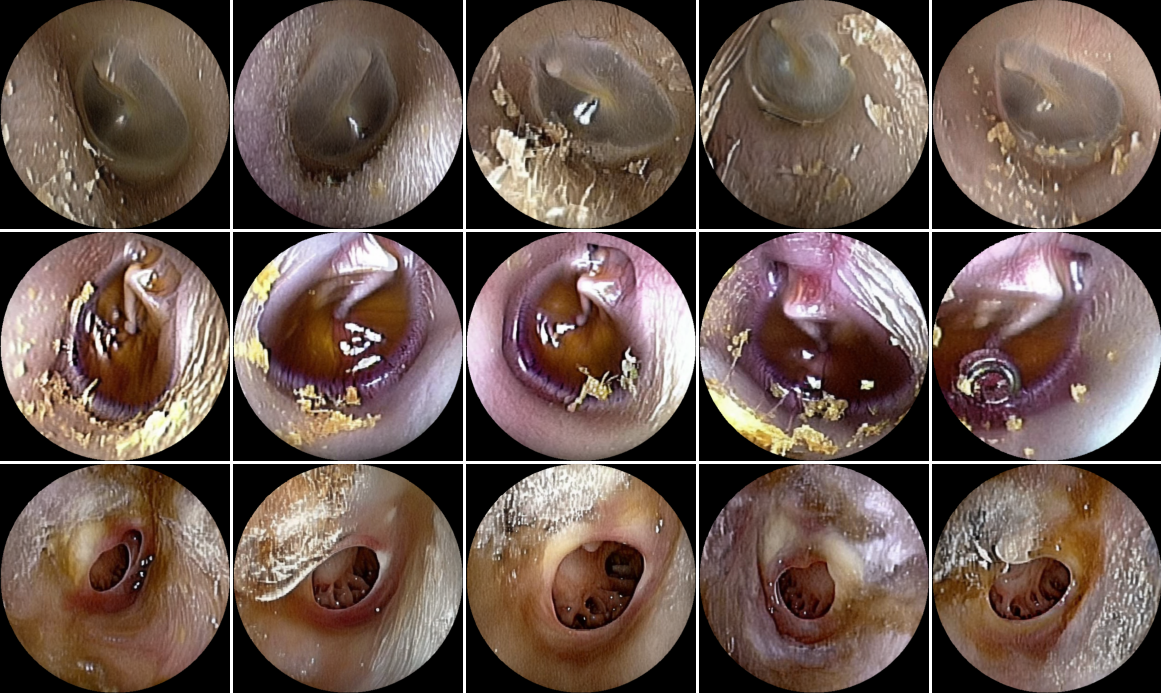}
  \caption{Samples of the CLUE synthetic dataset with $\sigma=0.0$.
  The rows correspond to: (1) Normal, (2) OME, (3) COM.
  }
  \label{fig:clue_sigma0p0}
\end{figure}

\clearpage
\subsection{CLUE ($\sigma=0.1$)}
\begin{figure}[ht]
  \centering
  \includegraphics[width=0.8\columnwidth]{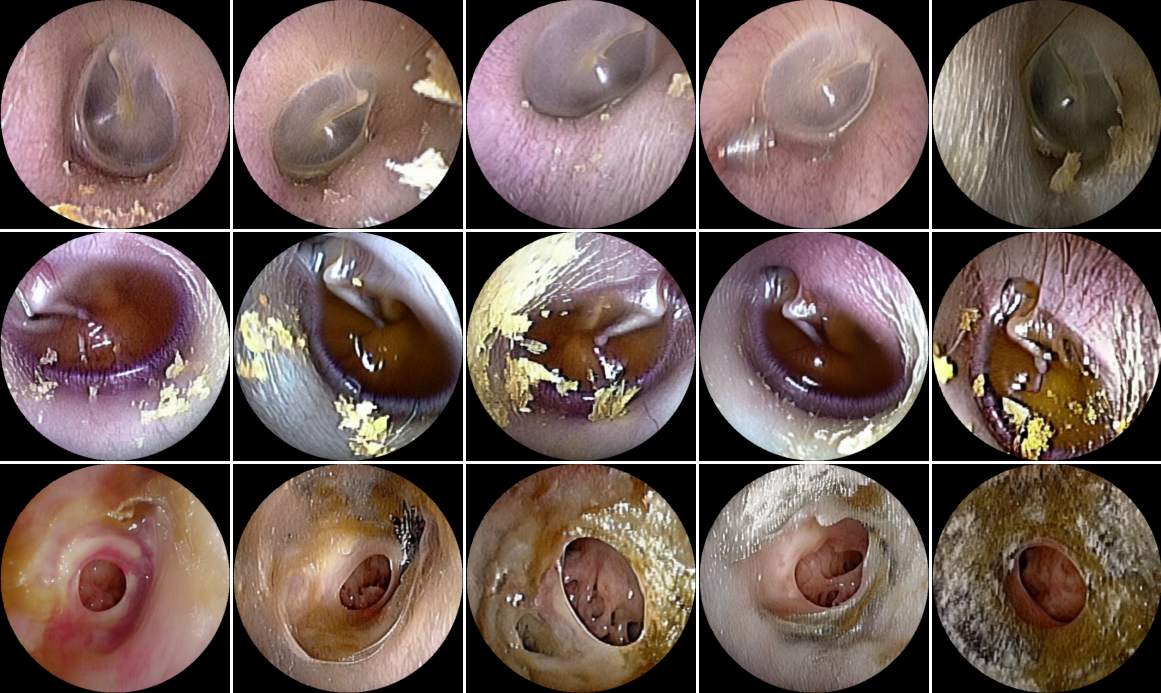}
  \caption{Samples of the CLUE synthetic dataset with $\sigma=0.1$.
  The rows correspond to: (1) Normal, (2) OME, (3) COM.
  }
  \label{fig:clue_sigma0p1}
\end{figure}

\subsection{CLUE ($\sigma=0.3$)}
\begin{figure}[ht]
  \centering
  \includegraphics[width=0.8\columnwidth]{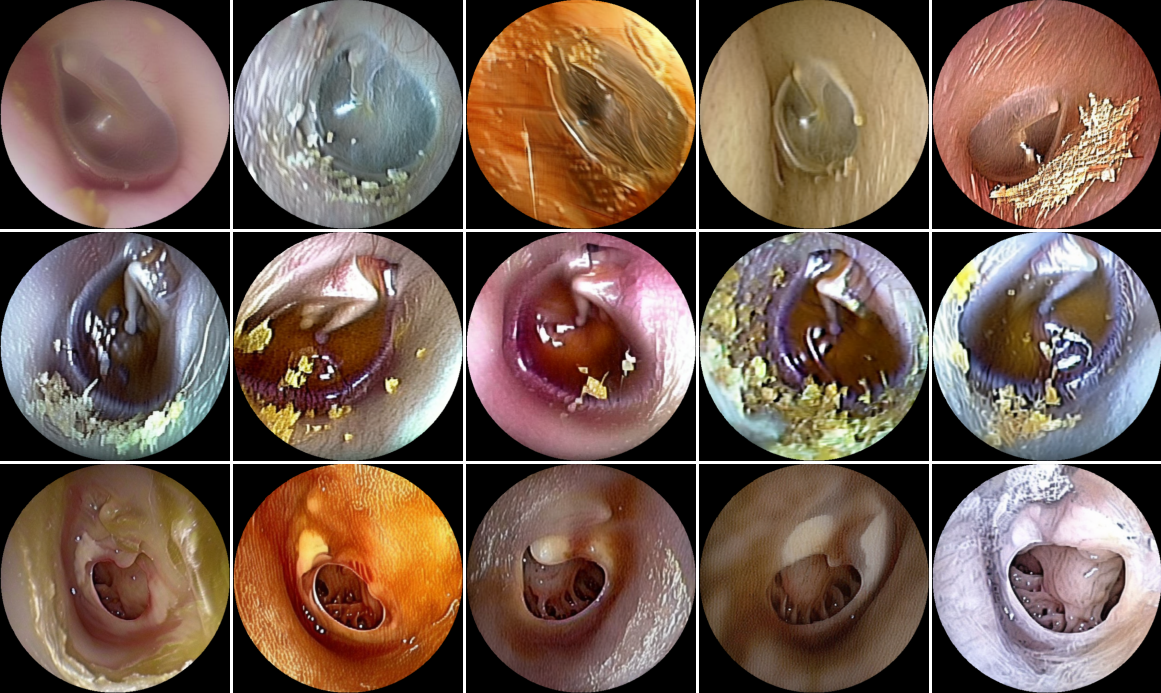}
  \caption{Samples of the CLUE synthetic dataset with $\sigma=0.3$.
  The rows correspond to: (1) Normal, (2) OME, (3) COM.
  }
  \label{fig:clue_sigma0p3}
\end{figure}

\clearpage

\subsection{CLUE ($\sigma=0.5$)}
\begin{figure}[ht]
  \centering
  \includegraphics[width=0.8\columnwidth]{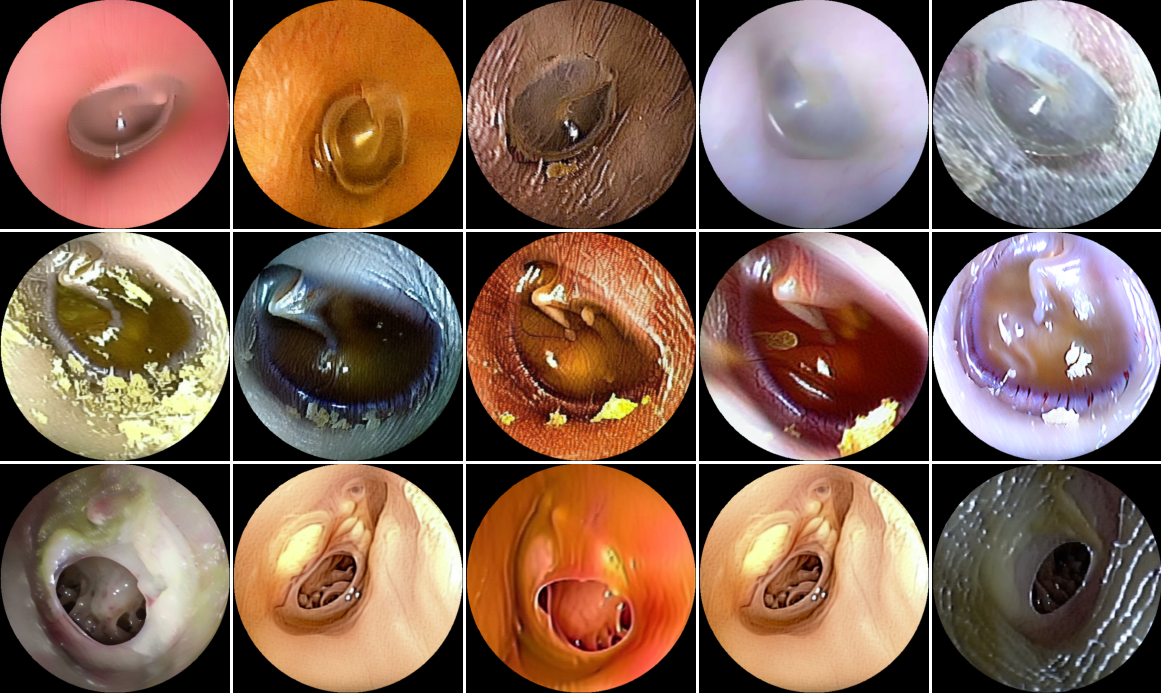}
  \caption{Samples of the CLUE synthetic dataset with $\sigma=0.5$.
  The rows correspond to: (1) Normal, (2) OME, (3) COM.
  }  \label{fig:clue_sigma0p5}
\end{figure}

\clearpage

\printcredits


\section*{Data availability}
This research (paper) used datasets from `The Open AI Dataset Project (AI-Hub, S. Korea)`. 
All data information can be accessed through `AI-Hub (\url{https://www.aihub.or.kr})`.


\bibliography{cas-refs}



\end{document}